\documentclass[10pt,twocolumn,letterpaper]{article}

\PassOptionsToPackage{table}{xcolor}
\usepackage{preprint}
\usepackage[T1]{fontenc}
\usepackage[utf8]{inputenc}
\usepackage{amsmath,amssymb,amsfonts}
\usepackage{graphicx}
\usepackage{array,booktabs,multirow,makecell,tabularx}
\usepackage{adjustbox}
\usepackage{threeparttable}
\usepackage{colortbl}
\usepackage{pifont}
\usepackage{textcomp}
\usepackage{listings}
\usepackage{microtype}
\usepackage{stfloats}
\usepackage{placeins}
\usepackage{etoolbox}

\definecolor{deepred}{RGB}{139,0,0}
\definecolor{deepgreen}{RGB}{0,100,0}
\providecommand{\best}[1]{\textbf{#1}}
\providecommand{\second}[1]{\underline{#1}}
\newcommand{\cmark}{\ding{51}}
\newcommand{\xmark}{\ding{55}}
\newcolumntype{Y}{>{\centering\arraybackslash}X}
\lstdefinelanguage{json}{
    morestring=[b]",
    morecomment=[l]{//},
    morecomment=[s]{/*}{*/}
}
\lstdefinestyle{jsonbox}{
    language=json,
    basicstyle=\ttfamily\scriptsize,
    frame=single,
    framerule=0.4pt,
    framesep=3pt,
    xleftmargin=0pt,
    xrightmargin=0pt,
    breaklines=true,
    breakatwhitespace=false,
    columns=fullflexible,
    keepspaces=true,
    showstringspaces=false,
    captionpos=b,
    aboveskip=8pt,
    belowskip=8pt
}

\AtBeginDocument{%
  \apptocmd{\normalsize}{%
    \setlength{\abovedisplayskip}{6pt plus 2pt minus 2pt}%
    \setlength{\belowdisplayskip}{6pt plus 2pt minus 2pt}%
    \setlength{\abovedisplayshortskip}{0pt plus 1pt}%
    \setlength{\belowdisplayshortskip}{4pt plus 1pt minus 1pt}%
  }{}{}%
  \normalsize
}

\definecolor{preprintblue}{rgb}{0.21,0.49,0.74}
\usepackage[breaklinks,colorlinks,allcolors=preprintblue]{hyperref}
\hypersetup{
  pdftitle={Count Anything},
  pdfauthor={Mengqi Lei, Shuokun Cheng, Wei Bao, Shaoyi Du, Jun-Hai Yong, Siqi Li, Yue Gao}
}

\title{Count Anything}
\author{%
  Mengqi Lei\textsuperscript{1}\quad
  Shuokun Cheng\textsuperscript{2}\quad
  Wei Bao\textsuperscript{1}\quad
  Shaoyi Du\textsuperscript{3} \\
  Jun-Hai Yong\textsuperscript{1}\quad
  Siqi Li\textsuperscript{1}\quad
  Yue Gao\textsuperscript{1} \\[4pt]
  {\small\begin{tabular}[t]{@{}c@{}}%
    \textsuperscript{1}\{BNRist, THUIBCS, School of Software\}, Tsinghua University \\
    \textsuperscript{2}School of Computer Science and Technology, China University of Geosciences, Wuhan \\
    \textsuperscript{3}State Key Laboratory of Human-Machine Hybrid Augmented Intelligence, \\
    National Engineering Research Center for Visual Information and Applications, \\
    and Institute of Artificial Intelligence and Robotics, Xi'an Jiaotong University%
  \end{tabular}}%
}

\begin{document}
\microtypesetup{activate=false}\raggedbottom
\maketitle
\addtocontents{toc}{\protect\setcounter{tocdepth}{-1}}

\begin{abstract}
Object counting remains largely fragmented across domain-specific datasets and task formulations, despite the rapid progress of generalist vision models. Existing counting models are often tailored to particular scenarios, such as crowds, vehicles, and cells, and therefore struggle to generalize across categories, visual domains, object scales, and density distributions. 
In this paper, we aim to count arbitrary targets specified by text across diverse visual domains, with every count supported by localized instances. This formulation unifies category-conditioned counting with interpretable spatial localization.
To support this goal, we construct \textbf{CLOC}, a \textbf{C}ross-domain \textbf{L}arge-scale \textbf{O}bject \textbf{C}ounting dataset that unifies heterogeneous annotations into a consistent counting format, comprising about 220K images, 619 categories, and 15M instances across six visual domains.
Based on CLOC, we propose \textbf{Count Anything}, a generalist model for cross-domain text-guided counting that predicts discrete instance points rather than density maps. To accommodate variations in object scale and density, it performs dual-granularity instance enumeration: a Region-level Sparse Counter provides object-level anchoring for large and sparse targets, while a Pixel-level Dense Counter captures small, crowded, and weakly bounded targets through dense point prediction.
A point-centric supervision strategy enables learning from heterogeneous annotations, and Complementary Count Fusion combines the outputs of both decoding branches based on their locations and confidence scores. 
Experiments on CLOC show that Count Anything reduces MAE from 23.38 (CountGD++) to 9.34 and achieves the best accuracy in all six domains, substantially outperforming existing open-world methods. Code and data are released at: \url{https://github.com/Mengqi-Lei/count-anything}.

\end{abstract}

\begin{figure}
    \centering
    \includegraphics[width=0.95\linewidth]{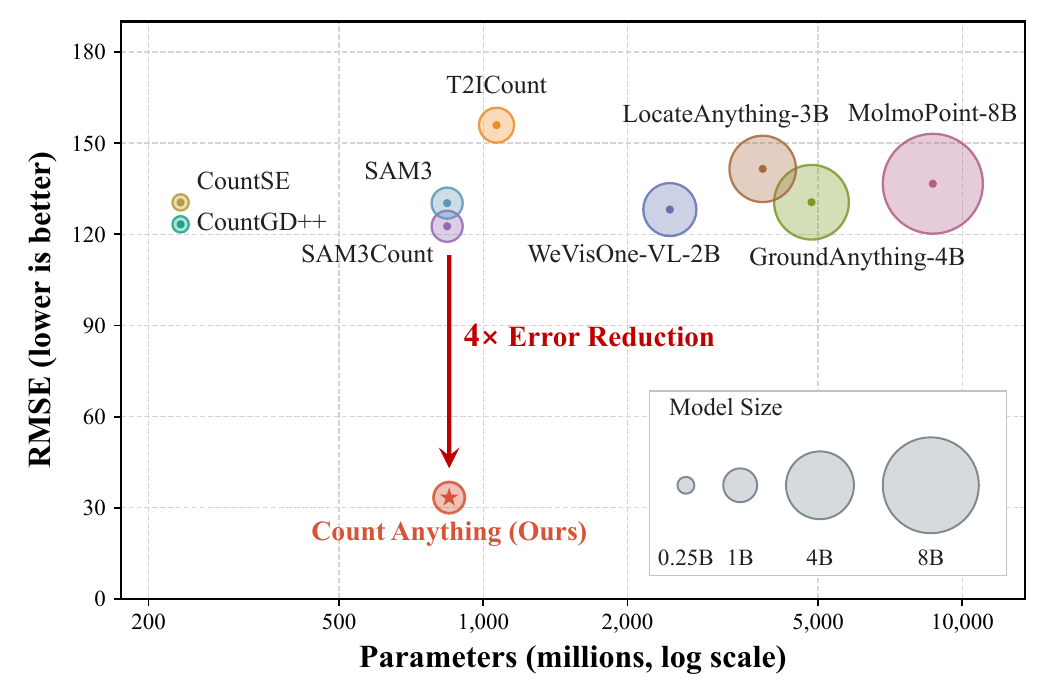}
    \vspace{-12pt}
    \caption{RMSE versus model size on CLOC. Count Anything's RMSE is only about $1/4$ of that of existing mainstream methods.}    \label{fig:comp_plot}
    \vspace{-12pt}
\end{figure}

\vspace{-10pt}
\section{Introduction}
\vspace{-6pt}
Vision foundation models are advancing visual understanding toward generalist systems that operate across categories, tasks, and visual domains~\cite{vfm_survey1,vfm_survey2,florence,phaselora,sam}.
Object counting, however, remains largely fragmented across domain-specific datasets~\cite{fsc133,fsc147,fscd_lvis,shanghaitech} and models~\cite{p2pnet,apgcc,cctrans}.
Existing systems are often developed for particular scenarios, such as crowds, vehicles, and cells, with limited coverage of target categories, imaging conditions, object scales, and density distributions.
A model that performs well on crowds may struggle with small vehicles in aerial imagery, densely packed nuclei in histopathology images, or repetitive crop organs in agricultural scenes.
These variations make it challenging to develop a single counting model that can serve diverse visual domains.

Recent text-guided and open-world counting methods have expanded counting beyond fixed target categories by drawing on vision-language representations and foundation models~\cite{countgd,countgdpp,clip_count,vlcounter}.
Text provides a convenient interface for specifying what to count, while instance-level predictions can provide spatial evidence for the resulting count.
Extending these capabilities across visual domains also requires broad training coverage and prediction mechanisms that accommodate substantial changes in object appearance, scale, and density.
In this work, we formulate generalist object counting as instance-grounded enumeration of user-specified targets across diverse visual domains.
Achieving this goal presents challenges in both data construction and model design.

On the data side, existing counting datasets typically cover individual application scenarios or a limited range of categories~\cite{fsc133,fsc147,fscd_lvis,shanghaitech}.
Other visual datasets contain rich instance annotations, but their annotation formats and definitions of countable instances differ.
Bounding boxes, points, masks, and polygons provide different forms of supervision~\cite{voc2007,mscoco,jhu_crowd,shanghaitech,fsc147,monuseg_dataset,nuinsseg,cellbindb}, while group annotations and ignored regions require additional handling to establish consistent counting targets~\cite{objects365,visdrone,monusac}.
A unified counting dataset must therefore ensure that each valid supervision unit corresponds to one independently countable instance.
To this end, we construct \textbf{CLOC}, a \textbf{C}ross-domain \textbf{L}arge-scale \textbf{O}bject \textbf{C}ounting dataset that unifies heterogeneous annotations from diverse visual sources into a consistent category-conditioned counting representation.
Each valid instance is associated with a counting point, with reliable bounding-box information retained when available.
CLOC comprises about 220K images, 619 categories, and 15M instances across six visual domains: General Scene, Remote Sensing, Histopathology, Cellular Microscopy, Agriculture, and Microbiology.
It provides a shared data foundation for training and evaluating generalist counting models.

On the model side, generalist counting must identify individual instances reliably across substantial variations in object scale, density, and boundary clarity.
Density-estimation methods support counting in crowded scenes~\cite{dmcount,clip_count,vlcounter}, but their continuous outputs do not directly establish correspondences between predictions and individual instances.
We therefore adopt discrete instance points as the final output representation, with the cardinality of the point set giving the count.
Reliable instance enumeration under this formulation requires object-level spatial constraints as well as fine-grained localization to distinguish neighboring instances.
Region-based counting uses object extent to provide geometric anchors for instance predictions~\cite{lsc_cnn,fscd_lvis,carpk}.
However, estimating reliable extents becomes difficult for small, crowded, or weakly bounded targets, potentially leading to missed or merged instances.
Dense point regression supports fine-grained localization without requiring complete object boundaries~\cite{p2pnet,apgcc,caapn}, but its lack of explicit object-level spatial constraints may lead to erroneous predictions for large or sparse targets.
These failure modes highlight the difficulty of using a single enumeration granularity across diverse counting scenarios.
We address this challenge through dual-granularity instance enumeration, integrating region-level geometric anchoring and pixel-level dense localization within a common point-set formulation.

Based on CLOC and this observation, we propose \textbf{Count Anything}, a generalist model for cross-domain text-guided counting.
A text-conditioned encoder extracts target-conditioned visual representations for two complementary decoding branches.
The Region-level Sparse Counter (RSC) predicts target regions and uses their centers as counting points, while the Pixel-level Dense Counter (PDC) generates dense point candidates from high-resolution features.
We design a point-centric supervision strategy to learn jointly from point-only and box-annotated instances.
During inference, Complementary Count Fusion (CCF) combines the branch outputs based on their spatial locations and confidence scores, reducing duplicate counts while preserving complementary predictions.
Together, these components produce a unified set of instance points for both counting and localization.
Extensive experiments on CLOC show that Count Anything achieves the best counting accuracy across all six visual domains, significantly outperforming existing methods.

Our contributions are summarized as follows:

\begin{itemize}

    \item We formulate generalist object counting around user-specified targets, instance-grounded predictions, and cross-domain coverage, providing a unified basis for dataset construction, model design, and evaluation.

    \item We construct \textbf{CLOC}, a large-scale cross-domain counting dataset that unifies heterogeneous instance annotations into a consistent category-conditioned counting representation across six visual domains.

    \item We propose \textbf{Count Anything}, a generalist model for text-guided counting that combines region-level object anchoring with dense point prediction through dual-granularity instance enumeration, supported by point-centric supervision and Complementary Count Fusion.

    \item Experiments on CLOC demonstrate that Count Anything achieves state-of-the-art counting performance across all six visual domains.
\end{itemize}

\vspace{-6pt}
\section{Related Work}
\vspace{-4pt}
\textbf{Object counting representations.}
Object counting methods employ density-, region-, and point-based prediction representations.
Density-estimation methods develop multi-scale features and point-based supervision to estimate counts in crowded scenes~\cite{csrnet,bl,dmcount,hyper_counter}.
Detection-based approaches provide explicit object extents and geometric structure for instance localization~\cite{lsc_cnn,carpk,fscd_lvis}.
Point-regression methods further connect counting with localization by predicting discrete instance points through one-to-one matching~\cite{p2pnet,apgcc}.
However, density-only outputs lack explicit instance correspondence; region-based predictions may miss dense targets, and point-only predictions lack explicit object-extent constraints.
Count Anything addresses these issues through dual-granularity instance enumeration, combining region-level anchoring with pixel-level dense localization in a common point-set formulation.

\textbf{Counting datasets and generalist counters.}
Domain-specific datasets establish counting benchmarks for crowds, vehicles, and microscopy~\cite{shanghaitech,jhu_crowd,carpk,bcdata}.
FSC-147, FSCD-LVIS, and OmniCount-191 broaden category coverage and support category-agnostic or multi-label counting~\cite{fsc147,fscd_lvis,omnicount191}.
Recent work further expands counting supervision: KubriCount and MixCount scale synthetic data for semantically fine-grained and mixed-object counting~\cite{kubricount,mixcount}, while PrACo++ and MUCCA evaluate semantic target selection under absent-category queries and distractors~\cite{praco_pp}.
However, scaling consistent instance supervision across heterogeneous real-world domains remains challenging because annotation forms and countable-instance definitions vary across sources.
CLOC addresses this challenge with about 220K images, 619 categories, and 15M instances across six visual domains, unifying heterogeneous annotations while covering broad count and density distributions from sparse to extremely dense scenes.

Generalist counters extend target coverage through transferable representations and semantic conditioning.
FamNet and CounTR develop category-agnostic counting representations~\cite{fsc147,countr}, while CLIP-Count, VLCounter, and T2ICount exploit vision-language representations or generative priors for text-guided counting~\cite{clip_count,vlcounter,t2icount}.
PseCo and the CountGD family further adapt foundation models to produce explicit instance candidates~\cite{pseco,countgd,countgdpp}.
SAM3Count and AdaCount improve dense-target recovery through adaptive region tiling and similarity-guided spatial and feature adaptation, respectively~\cite{sam3count,adacount}.
However, jointly maintaining semantic target selection and reliable instance enumeration across heterogeneous domains, object scales, and densities remains challenging.
Count Anything tackles this challenge within a unified framework trained on CLOC, combining dual-granularity enumeration, point-centric supervision, and Complementary Count Fusion.
It achieves an overall MAE of 9.34 and the lowest MAE across all six CLOC domains among the evaluated methods.
An extended discussion of related work is provided in Appendix~\ref{sec:extended_related_work}.

\vspace{-6pt}
\section{CLOC Dataset}
\label{sec:cloc_dataset}
\vspace{-6pt}
To support cross-domain text-guided object counting, we construct \textbf{CLOC}, a \textbf{C}ross-domain \textbf{L}arge-scale \textbf{O}bject \textbf{C}ounting dataset for generalist text-guided counting across domains. Existing counting datasets are mostly built around specific scenarios or target types, such as crowds, vehicles, cells, or crops~\cite{shanghaitech,carpk,bcdata,mtc}, and thus provide limited support for training and evaluating generalist counting models across visual domains, target categories, object scales, and density distributions. Meanwhile, existing large-scale detection, segmentation, and visual grounding datasets provide rich region- or instance-level annotations~\cite{mscoco,objects365,lvis,flickr30k_entities}, but their original task objectives and annotation protocols do not directly satisfy the one-instance-one-count supervision required by counting.

To this end, CLOC reorganizes images, categories, and instance annotations from multiple public datasets around a text-guided counting task. In CLOC, category names are used as textual queries, so the category-specified counting protocol is treated as a concrete form of text-guided counting: given an image and a text query, the model is required to predict the number of instances corresponding to the queried target in the image. CLOC covers six visual domains: General Scene, Remote Sensing, Histopathology, Cellular Microscopy, Agriculture, and Microbiology, and contains about $220$K images, $619$ categories, and $15.4$M instances. Compared with existing counting datasets, CLOC provides both a larger scale and broader cross-domain coverage, while preserving long-tailed category distributions and target-count variations from low-density to extremely dense scenes. It therefore provides a unified benchmark for training and evaluating generalist object counting models. More detailed dataset statistics and the construction pipeline are provided in Appendix~\ref{app:cloc_details}.

\begin{figure*}[t]
    \centering
    \includegraphics[width=0.95\linewidth]{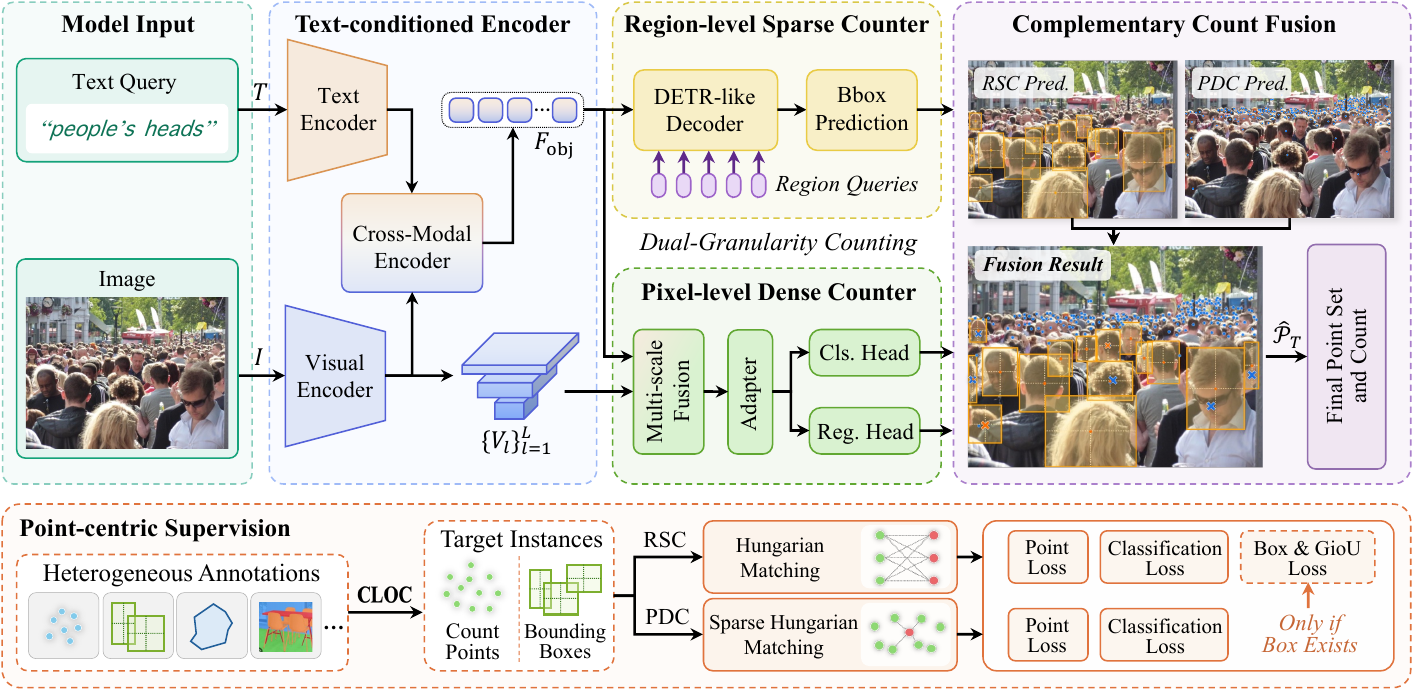}
    \vspace{-4pt}
\caption{Overview of Count Anything. Text-conditioned features drive Region-level Sparse Counter (RSC) and Pixel-level Dense Counter (PDC) predictions, which Complementary Count Fusion (CCF) merges into the final counting point set. The bottom panel shows point-centric supervision: both branches learn from instance points, while reliable boxes provide additional geometric supervision for RSC.}
    \label{fig:countanything}
    \vspace{-12pt}
\end{figure*}

\vspace{-4pt}
\section{Count Anything Model}
\vspace{-2pt}
\label{sec:method}

\subsection{Method Overview}
\label{overview}
\vspace{-4pt}

Given an image $I$ and a text query $T$, Count Anything predicts the visible target instances as a point set:
\begin{equation}
\hat{\mathcal{P}}_T
=\{(\hat{p}_n,\hat{s}_n)\}_{n=1}^{\hat{N}},
\quad \hat{p}_n\in\mathbb{R}^{2},\quad \hat{s}_n\in[0,1],
\end{equation}
where $\hat{p}_n$ and $\hat{s}_n$ are the location and confidence of the $n$-th instance.
The count is $\hat{c}_T=|\hat{\mathcal{P}}_T|$, with each predicted instance providing spatial evidence.

As shown in Fig.~\ref{fig:countanything}, Count Anything combines dual-granularity instance prediction with complementary fusion:
\begin{equation}
\label{eq:countanything_pipeline}
\setlength{\jot}{0pt}
\begin{aligned}
&(I,T)
\xrightarrow{\Phi_{\mathrm{enc}}}
\left(F_{\mathrm{obj}},\{V_l\}_{l=1}^{L}\right)\\
&\quad\xrightarrow{}
\left\{
\begin{aligned}
\mathcal{R}&=\operatorname{RSC}(F_{\mathrm{obj}}),\\
\mathcal{D}&=\operatorname{PDC}(F_{\mathrm{obj}},\{V_l\}_{l=1}^{L}),
\end{aligned}
\right.
\xrightarrow{\operatorname{CCF}}
\hat{\mathcal{P}}_T.
\end{aligned}
\end{equation}
Here, $\Phi_{\mathrm{enc}}$ produces the target-conditioned representation $F_{\mathrm{obj}}$ and multi-scale visual features $\{V_l\}_{l=1}^{L}$; $\mathcal{R}$ and $\mathcal{D}$ denote the region and dense point candidates, respectively.
The Region-level Sparse Counter (RSC) predicts object regions and uses their centers as counting points.
The Pixel-level Dense Counter (PDC) combines target-conditioned and high-resolution features to localize small, crowded, and weakly bounded instances.
Both counters learn from counting points, with reliable bounding boxes providing additional geometric supervision for RSC.
During inference, Complementary Count Fusion (CCF) reconciles their predictions into the final instance point set.

\vspace{-2pt}
\subsection{Dual-Granularity Instance Enumeration}
\vspace{-2pt}
\textbf{Text-conditioned encoding.}
We adopt the image-language encoding backbone of pretrained SAM3~\cite{sam3} as the text-conditioned encoder.
The visual and text encoders produce $\{V_l\}_{l=1}^{L}=E_v(I)$ and $e_T=E_t(T)$, respectively.
A cross-modal encoder then obtains the target-conditioned representation
$F_{\mathrm{obj}}=\Phi_{\mathrm{obj}}(\{V_l\}_{l=1}^{L},e_T)$.
The encoder adaptation and training setting are described in Sec.~\ref{sec:implementation_details}.

\textbf{Region-level sparse enumeration.}
RSC uses a DETR-like decoder~\cite{detr} with $Q_r$ region queries.
For query $i$, it predicts a foreground logit and bounding box,
$(z_i^r,\hat{B}_i^r)=R_i(F_{\mathrm{obj}})$.
The counting point is $\hat{p}_i^r=\operatorname{center}(\hat{B}_i^r)$ and its confidence is $\hat{s}_i^r=\sigma(z_i^r)$, where $\sigma$ is the sigmoid function.
The resulting candidate set is $\mathcal{R}=\{(\hat{B}_i^r,\hat{p}_i^r,\hat{s}_i^r)\}_{i=1}^{Q_r}$.
Region predictions retain object extent and provide geometric anchors for instance enumeration.

\textbf{Pixel-level dense enumeration.}
Small, crowded, and weakly bounded targets require fine spatial detail to distinguish individual instances.
Although $F_{\mathrm{obj}}$ provides target-conditioned information for object-level reasoning, its limited spatial resolution can restrict dense-instance localization.
PDC therefore constructs a dense counting representation that combines semantic target information with local spatial detail:
\begin{equation}
F_{\mathrm{pix}}=\Phi_{\mathrm{pix}}\left(
\operatorname{Align}(F_{\mathrm{obj}}),
\{\operatorname{Align}(V_l)\}_{l\in\mathcal{S}}
\right),
\end{equation}
where $\mathcal{S}\subseteq\{1,\ldots,L\}$ selects high-resolution visual features, $\operatorname{Align}$ maps features onto a common spatial grid, and $\Phi_{\mathrm{pix}}$ performs lightweight convolutional fusion.
The fused representation carries query-dependent information from $F_{\mathrm{obj}}$ together with the finer spatial cues in $V_l$, supporting target selection and the separation of neighboring instances on the same dense grid.
To adapt these fused features to instance enumeration, we introduce a residual-block adapter,
$G_{\mathrm{pix}}=A_{\mathrm{pdc}}(F_{\mathrm{pix}})$.
The adapter preserves the spatial grid while learning counting-specific local representations for query-conditioned foreground discrimination and point localization.

Inspired by the point-regression formulation of P2PNet \cite{p2pnet}, PDC enumerates candidates on a dense grid $\Omega=\{u_k\}_{k=1}^{Q_d}$, with an image-space anchor $a_k=\pi(u_k)$ for each location.
The grid provides dense candidate coverage, while offset regression refines each candidate relative to its anchor.
Given the local feature $g_k=G_{\mathrm{pix}}(u_k)$, classification and regression heads predict:
\begin{equation}
\begin{aligned}
z_k^d&=D_{\mathrm{cls}}(g_k),
&\Delta_k&=D_{\mathrm{reg}}(g_k),\\
\hat{p}_k^d&=a_k+\rho\Delta_k,
&\hat{s}_k^d&=\operatorname{softmax}(z_k^d)_{\mathrm{fg}}.
\end{aligned}
\end{equation}
Here, $z_k^d\in\mathbb{R}^2$ contains background-foreground logits, $\Delta_k\in\mathbb{R}^2$ is the coordinate offset, and $\rho$ scales the offset.
The resulting candidates $\mathcal{D}=\{(\hat{p}_k^d,\hat{s}_k^d)\}_{k=1}^{Q_d}$ provide discrete locations and foreground confidence scores for subsequent point-level matching and complementary fusion.

\vspace{-4pt}

\subsection{Point-centric Supervision}
\vspace{-4pt}

Point-centric supervision uses instance locations as a common training signal and incorporates reliable object extents when available.
For a training sample, query-associated annotations are
$\mathcal{Y}_T=\{(p_j,B_j,m_j)\}_{j=1}^{N}$, where $p_j$ is a counting point, $B_j$ is an optional box, and $m_j\in\{0,1\}$ indicates box availability.
The counting point is the box center when $m_j=1$, and the provided point annotation otherwise.
Both annotation types therefore support instance matching and localization, while box regression is restricted to instances with reliable box annotations.

\textbf{Branch-specific matching.}
Both counters use one-to-one assignment based on point distance and foreground confidence.
For RSC, point-based matching accommodates both box-annotated and point-only instances:
\begin{equation}
C^r_{ij}=\lambda_p^r\,d(\hat{p}_i^r,p_j)-\lambda_s^r\,\hat{s}_i^r,
\end{equation}
where $d$ is the point-distance term and the weights balance localization and confidence.
Hungarian matching gives $\mathcal{M}_r=\operatorname{Hungarian}(C^r)$.
PDC produces a much larger candidate set, making a full candidate--instance cost matrix expensive.
We therefore restrict matching to local candidates $\mathcal{N}(j)$ around each ground-truth point:
\begin{equation}
C^d_{kj}=\lambda_p^d\|\hat{p}_k^d-p_j^{\mathrm{img}}\|_2
-\lambda_s^d\hat{s}_k^d,\quad k\in\mathcal{N}(j),
\end{equation}
where $p_j^{\mathrm{img}}$ is the target in input-image coordinates.
The assignment $\mathcal{M}_d=\operatorname{SparseHungarian}(C^d)$ retains one-to-one matching while reducing the number of candidate--instance edges considered.

\textbf{RSC supervision.}
All matched RSC predictions receive point localization supervision:
\begin{equation}
\mathcal{L}_{point}^{r}=\frac{1}{Z_r}
\sum_{(i,j)\in\mathcal{M}_r}\|\hat{p}_i^r-p_j\|_2^2,
\end{equation}
where $Z_r$ is the number of ground-truth instances in the batch.
Only reliable ground-truth boxes contribute to geometric regression:
\begin{equation}
\begin{aligned}
\mathcal{L}_{box}^{r}&=\frac{1}{Z_b}
\sum_{(i,j)\in\mathcal{M}_r}m_j\|\hat{B}_i^r-B_j\|_1,\\
\mathcal{L}_{giou}^{r}&=\frac{1}{Z_b}
\sum_{(i,j)\in\mathcal{M}_r}m_j
\left(1-\operatorname{GIoU}(\hat{B}_i^r,B_j)\right),
\end{aligned}
\end{equation}
with $Z_b$ normalizing valid box-annotated instances.

Matched regions can differ in spatial coverage, so RSC uses soft foreground targets that reflect coverage quality as well as predicted confidence.
For each matched pair, define
\[
B_{ij}^{\star}=\begin{cases}
B_j,&m_j=1,\\
\tilde{B}_{ij},&m_j=0,
\end{cases}
\]
where $\tilde{B}_{ij}$ is centered at $p_j$ and has the size of the predicted box $\hat{B}_i^r$.
For point-only instances, this auxiliary region assesses alignment at the predicted scale; it is used solely for classification quality estimation, never as a box regression target.
Coverage quality and the soft target are
\begin{equation}
\begin{aligned}
g_{ij}&=\frac{|\hat{B}_i^r\cap B_{ij}^{\star}|}{|B_{ij}^{\star}|},\\
\tilde{y}_i&=\max\left(\epsilon,
\operatorname{sg}(\hat{s}_i^r)^{\alpha}(g_{ij})^{1-\alpha}\right),
\end{aligned}
\end{equation}
where $\operatorname{sg}$ denotes stop-gradient, $\epsilon$ is the minimum positive label, and $\alpha$ balances confidence and coverage.
Unmatched queries receive background labels, and classification uses
$\mathcal{L}_{cls}^{r}=\mathcal{L}_{\mathrm{BCE}}(\{z_i^r\},\{\tilde{y}_i\})$.

\textbf{PDC supervision.}
Matched dense candidates are foreground and the remaining candidates are background.
PDC uses $\mathcal{L}_{cls}^{d}=\mathcal{L}_{\mathrm{CE}}(\{z_k^d\},\{y_k^d\})$, where $y_k^d\in\{\mathrm{bg},\mathrm{fg}\}$, together with
\vspace{-4pt}
\begin{equation}
\mathcal{L}_{point}^{d}=\frac{1}{Z_d}
\sum_{(k,j)\in\mathcal{M}_d}\|\hat{p}_k^d-p_j^{\mathrm{img}}\|_2^2,
\vspace{-4pt}
\end{equation}
where $Z_d$ normalizes matched foreground samples.

The overall objective is
\begin{equation}
\begin{aligned}
\mathcal{L} ={}&
\lambda_{cls}^{r}\mathcal{L}_{cls}^{r}
+ \lambda_{point}^{r}\mathcal{L}_{point}^{r}
+ \lambda_{box}^{r}\mathcal{L}_{box}^{r} \\
&+ \lambda_{giou}^{r}\mathcal{L}_{giou}^{r}
+ \lambda_{cls}^{d}\mathcal{L}_{cls}^{d}
+ \lambda_{point}^{d}\mathcal{L}_{point}^{d},
\end{aligned}
\end{equation}
where each $\lambda$ is a loss weight term.
Both counters thus learn instance presence and location, with explicit box regression applied only where the annotations support geometric supervision.

\vspace{-4pt}
\subsection{Complementary Count Fusion}
\vspace{-4pt}

\begin{figure}[t]
    \centering
    \includegraphics[width=0.75\linewidth]{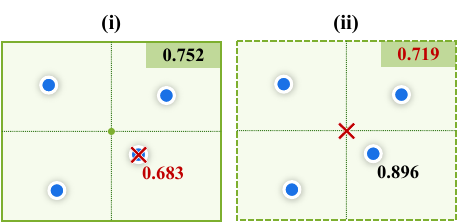}
    \vspace{-6pt}
    \caption{CCF retains the higher-confidence prediction between an RSC center and its nearest enclosed PDC point; other PDC points remain unchanged by this comparison.}
    \label{fig:ccf}
    \vspace{-14pt}
\end{figure}

During inference, RSC and PDC produce region-level and dense point candidates for the same queried target.
Direct concatenation may double-count instances predicted by both branches.
Complementary Count Fusion (CCF) combines the two candidate sets through local spatial association and confidence-based selection, reducing duplicate counts while retaining complementary predictions.

We first filter RSC predictions by the confidence threshold $\tau_r$ and perform region-level suppression using intersection over minimum area (IoM),
$\operatorname{IoM}(A,B)=|A\cap B|/\min(|A|,|B|)$.
If one region contains another or their IoM exceeds $\tau_{\mathrm{IoM}}$, only the higher-confidence region is retained.
PDC candidates are filtered by threshold $\tau_d$.

For cross-branch fusion, each retained RSC candidate contributes one counting hypothesis at its region center. CCF uses the predicted region to restrict the search for a potential duplicate. If no PDC point lies inside the region, the RSC center is retained. Otherwise, the enclosed PDC point nearest to the RSC center is selected as a potential duplicate, and the higher-scoring prediction of the pair is retained (Fig.~\ref{fig:ccf}). To preserve multiple dense predictions within the same region, CCF limits each comparison to one candidate pair, leaving the other PDC points unchanged by this comparison.

Let $\mathcal{P}^{r}_{\mathrm{keep}}$ and $\mathcal{P}^{d}_{\mathrm{keep}}$ denote the surviving RSC and PDC points, respectively.
The final point set and count are
$\hat{\mathcal{P}}_T=\mathcal{P}^{r}_{\mathrm{keep}}\cup\mathcal{P}^{d}_{\mathrm{keep}}$
and $\hat{c}_T=|\hat{\mathcal{P}}_T|$.
This local selection directly uses the confidence estimates learned by the two counters and introduces no additional trainable fusion parameters.

\vspace{-6pt}
\section{Experiments}
\vspace{-4pt}

\begin{table*}[t]
\centering
\scriptsize
\setlength{\tabcolsep}{4pt}
\renewcommand{\arraystretch}{0.9}
\captionsetup{position=top,skip=1pt}
\caption{Comparison with other state-of-the-art methods on the CLOC dataset. The first block lists object counting methods, and the second lists general-purpose detection, segmentation, and visual grounding models.}
\label{tab:merged_full_domain_mae}
\begin{tabular*}{\linewidth}{@{\extracolsep{\fill}}llccccccccc@{}}
\toprule

\multirow{2}{*}{\textbf{Method}} &
\multirow{2}{*}{\textbf{Venue}} &
\multirow{2}{*}{\textbf{Ins.Grd.}} &
\multicolumn{2}{c}{\textbf{Full Test}} &
\multicolumn{6}{c}{\textbf{Single Domain MAE}} \\
\cmidrule(lr){4-5}\cmidrule(lr){6-11}
 & & &
\textbf{MAE} & \textbf{RMSE} &
\textbf{GS} & \textbf{RS} & \textbf{Histo.} & \textbf{Cell.} & \textbf{Agri.} & \textbf{Micro.} \\
\midrule

 LOCA~\cite{loca} & ICCV'23 & \xmark & 68.86 & 179.26 & 55.24 & 81.37 & 65.86 & 124.36 & 473.81 & 39.47 \\
 CLIP-Count~\cite{clip_count} & ACMMM'23 & \xmark & 31.28 & \second{118.55} & 24.07 & 25.81 & 57.98 & 121.17 & 298.45 & 22.40 \\
 CounTX~\cite{countx} & BMVC'23 & \xmark & 32.24 & 122.10 & 23.84 & 30.06 & 64.64 & \second{112.22} & 301.10 & 27.41 \\
 VLCounter~\cite{vlcounter} & AAAI'24 & \xmark & 30.52 & 121.88 & 23.59 & 22.45 & 55.14 & 125.83 & 306.38 & 37.72 \\
 VA-Count~\cite{va_count} & ECCV'24 & \xmark & 33.85 & 129.81 & 22.39 & 36.62 & 72.50 & 163.73 & 315.23 & 20.47 \\
 CountGD~\cite{countgd} & NeurIPS'24 & \cmark & 55.91 & 1081.82 & 53.83 & 42.23 & 99.69 & 114.50 & \second{244.80} & 11.09 \\
 YOLO-Count~\cite{yolo_count} & ICCV'25 & \xmark & 34.15 & 133.56 & 25.62 & 27.69 & 72.50 & 168.37 & 309.30 & 24.33 \\
 T2ICount~\cite{t2icount} & CVPR'25 & \xmark & 50.02 & 155.92 & 34.08 & 62.46 & 73.06 & 151.30 & 296.37 & 203.51 \\
 CountSE~\cite{countse} & ICCV'25 & \xmark & 28.81 & 130.49 & 22.14 & 19.07 & 70.27 & 120.08 & 302.19 & 22.22 \\
 CountGD++~\cite{countgdpp} & CVPR'26 & \cmark & 23.38 & 123.28 & \second{14.50} & 10.69 & 55.83 & 222.25 & 321.99 & 43.80 \\
 SAM3Count~\cite{sam3count} & CVPRW'26 & \cmark & 22.84 & 122.64 & 16.20 & 10.25 & 63.11 & 128.50 & 307.30 & 34.44 \\

\midrule

 GLIP-L~\cite{glip} & CVPR'22 & \cmark & 37.38 & 140.07 & 31.06 & 19.57 & 68.78 & 225.42 & 320.85 & 43.71 \\
 YOLO-World-X~\cite{yolo_world} & CVPR'24 & \cmark & 34.94 & 138.35 & 28.17 & 17.81 & 67.93 & 226.07 & 321.13 & 42.49 \\
 GrdDINO-SwinB~\cite{grounding_dino} & ECCV'24 & \cmark & 35.06 & 139.46 & 29.50 & 14.13 & 69.11 & 224.96 & 320.17 & 43.75 \\
 YOLOE-v8L~\cite{yoloe} & ICCV'25 & \cmark & 40.03 & 140.62 & 33.97 & 21.78 & 69.99 & 228.28 & 321.97 & 44.80 \\
 YOLOE-11L~\cite{yoloe} & ICCV'25 & \cmark & 40.05 & 140.60 & 34.00 & 21.78 & 69.95 & 228.24 & 321.88 & 44.81 \\
 SAM3~\cite{sam3} & ICLR'26 & \cmark & 25.80 & 130.26 & 19.39 & 10.88 & 64.32 & 152.45 & 315.88 & 37.04 \\
 GroundAnything-4B~\cite{groundanything} & arXiv'26 & \cmark & 25.67 & 130.56 & 19.51 & 11.56 & 63.24 & 163.01 & 299.05 & 19.78 \\
 WeVisOne-VL-2B~\cite{wevisone} & arXiv'26 & \cmark & \second{22.51} & 128.14 & 17.52 & \second{7.88} & \second{43.72} & 163.18 & 287.70 & \second{7.16} \\
MolmoPoint-8B~\cite{molmopoint} & ECCV'26 & \cmark & 29.29 & 136.64 & 23.49 & 11.46 & 63.06 & 217.51 & 300.01 & 19.56 \\
LocateAnything-3B~\cite{locateanything} & ECCV'26 & \cmark & 31.77 & 141.52 & 26.27 & 12.35 & 76.00 & 198.35 & 305.21 & 31.18 \\

\midrule

\textbf{Count Anything} & -- & \cmark & \best{9.34} & \best{33.34} & \best{8.76} & \best{6.21} & \best{16.64} & \best{38.65} & \best{41.78} & \best{4.30} \\

\bottomrule
\end{tabular*}
\par\vspace{2pt}
{\scriptsize\raggedright
\noindent\textit{Note:} Ins.Grd. denotes instance grounding capability.
GS, RS, Histo., Cell., Agri., and Micro. denote General Scene, Remote Sensing, Histopathology, Cellular Microscopy, Agriculture, and Microbiology, respectively.
\par}
\vspace{-14pt}
\end{table*}

\subsection{Experimental Setup}
\label{sec:implementation_details}
\vspace{-4pt}
Count Anything adopts the pretrained SAM3~\cite{sam3} text-conditioned encoder with its original weights frozen, and introduces LoRA~\cite{lora} adapters into the cross-modal encoder for counting-task adaptation. All experiments are conducted on a single node with $8$ NVIDIA A100 GPUs. We train the model for $30$ epochs with an input resolution of $1008\times1008$ and a per-GPU batch size of $24$. The learning rates of RSC, PDC, and LoRA are set to $1\times10^{-5}$, $1\times10^{-4}$, and $1\times10^{-3}$, respectively, and AdamW~\cite{adamw} is used as the optimizer. RSC uses $200$ region queries. During inference, the score thresholds are $\tau_r=0.5$ and $\tau_d=0.5$. The IoM threshold $\tau_{\mathrm{IoM}}$ for RSC duplicate removal is set to $0.5$. We report results on the CLOC test split. Following prior object-counting studies, we report MAE and RMSE in the main paper. We additionally report $\mathrm{nAP}_{25}$, $\mathrm{nAP}_{50}$, and $\mathrm{nAP}_{05\,:\,50}$ \cite{p2pnet} to evaluate instance localization performance. More implementation and evaluation details are provided in Appendix~\ref{app:implementation_details}.

\vspace{-4pt}
\subsection{Comparison with Other Methods}
\label{sec:comparison_with_other_methods}
\textbf{Main results on CLOC.}
We compare Count Anything with representative open-world object counting methods, as well as detection, segmentation, and visual grounding models. For all competing methods, we use their officially released checkpoints and standard inference pipelines to evaluate their out-of-the-box capability. Under this setting, we aim to answer a practical question: \emph{How far can current counting models go when facing larger-scale data, more diverse visual scenarios, and broader count distributions?}

\begin{figure*}[t]
    \centering
    \includegraphics[width=0.95\textwidth]{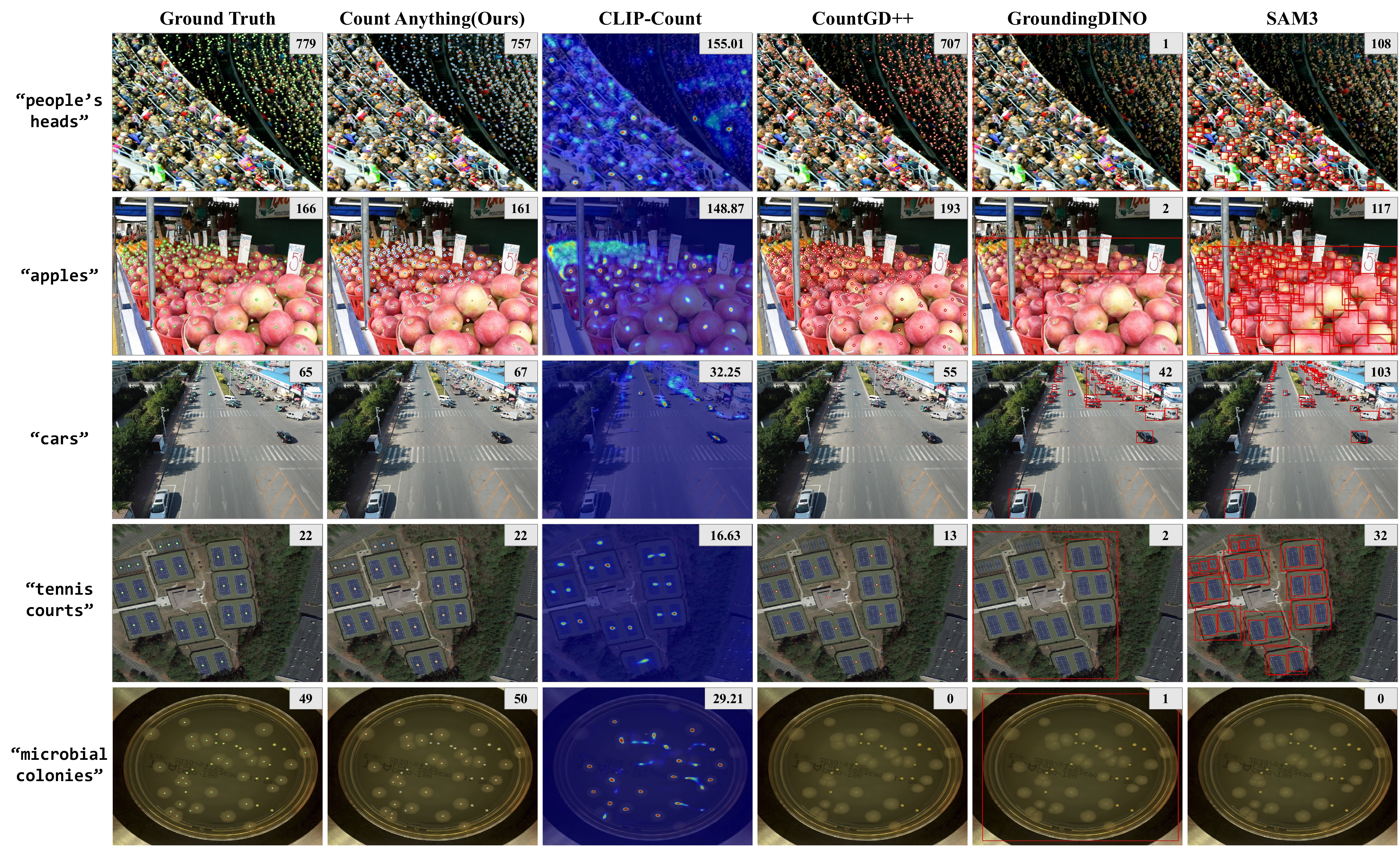}
    \vspace{-8pt}
\caption{Qualitative comparison on CLOC with varying scenes and domains. Rows correspond to the text queries on the left; columns show ground truth and model outputs. Numbers denote counts, with points, density maps, or boxes visualizing different models' outputs.}    \label{fig:quantitative_visual_comparison}
    \vspace{-14pt}
\end{figure*}

As shown in Table~\ref{tab:merged_full_domain_mae} and Figure~\ref{fig:comp_plot}, Count Anything achieves 9.34 MAE and 33.34 RMSE on CLOC, substantially outperforming all compared methods across the six visual domains. The results show that existing open-world object counting methods are effective to some extent in the general scene, but their errors increase noticeably in specialized domains such as histopathology, cellular microscopy, agriculture, and microbiology. The evaluated detection, segmentation, and visual grounding models can produce instance-level predictions, but they are still prone to missed detections or merged instances in scenes with dense, small, or weakly bounded targets, making it difficult to provide reliable counting results. In contrast, Count Anything performs more stably in the cross-domain counting task, demonstrating better generalization ability.
Figure~\ref{fig:quantitative_visual_comparison} shows visualized predictions of different models on several samples. Compared with these methods, Count Anything's predictions are noticeably closer to the ground truth across different scenarios. In particular, Count Anything maintains stable counting results on samples with significant differences in target scale and density distribution.

We attribute the substantial advantage of Count Anything to two key factors: first, the dual-granularity counting design that combines region-level sparse counting and pixel-level dense counting in the model architecture; second, the larger-scale and broader coverage of categories, visual domains, and target-density distributions provided by CLOC. We further disentangle and validate the contributions of these two factors in the following ablation studies.

\begin{table}[t]
    \centering
    \scriptsize
    \setlength{\tabcolsep}{2.2pt}
    \renewcommand{\arraystretch}{0.8}
    \captionsetup{position=top,skip=1pt}
    \caption{Comparison of instance localization performance.}
    \label{tab:comp_localization}
    \begin{tabular*}{\columnwidth}{@{\extracolsep{\fill}}lccc@{}}
        \toprule
        \textbf{Method} & $\mathbf{nAP}_{25}$ & $\mathbf{nAP}_{50}$ & $\mathbf{nAP}_{05\,:\,50}$ \\
        \midrule
        CountGD~\cite{countgd} & 30.35 & 40.55 & 27.90 \\
        CountGD++~\cite{countgdpp} & 35.86 & 45.43 & 32.09 \\
        SAM3~\cite{sam3} & 23.63 & 26.42 & 21.34 \\
        LocateAnything-3B~\cite{locateanything} & 12.15 & 15.75 & 11.31 \\
        \midrule
        \textbf{Count Anything} & \textbf{53.93} & \textbf{70.27} & \textbf{50.26} \\
        \bottomrule
    \end{tabular*}
    \vspace{-14pt}
\end{table}

\textbf{Instance localization performance.}
To further evaluate instance localization performance, we compare Count Anything with representative counting and visual grounding methods. As shown in Table~\ref{tab:comp_localization}, Count Anything achieves the best results on $\mathrm{nAP}_{25}$, $\mathrm{nAP}_{50}$, and $\mathrm{nAP}_{05\,:\,50}$, showing that it can more accurately localize target instances corresponding to the text query. Together with the counting results, this comparison shows that Count Anything achieves strong counting accuracy while also providing accurate instance localization.

\begin{table}[t]
    \centering
    \scriptsize
    \setlength{\tabcolsep}{1.6pt}
    \renewcommand{\arraystretch}{0.8}
    \captionsetup{position=top,skip=1pt}
    \caption{Comparison on the SHTech Part-A subset of CLOC test set, including foundation models and specialized crowd counters.}
    \label{tab:comp_crowd}
    \begin{tabular*}{\columnwidth}{@{\extracolsep{\fill}}llcc@{}}
    \toprule
    \textbf{Method} & \textbf{Venue} & \textbf{MAE} & \textbf{RMSE} \\
    \midrule
    GrdDINO-SwinB~\cite{grounding_dino} & ECCV'24 & 450.86 & 588.65 \\
    YOLO-World-X~\cite{yolo_world} & CVPR'24 & 421.15 & 571.57 \\
    SAM3~\cite{sam3} & ICLR'26 & 345.55 & 527.10 \\
    \midrule
    APGCC~\cite{apgcc} & ECCV'24 & \second{49.30} & \best{75.03} \\
    P2PNet~\cite{p2pnet} & ICCV'21 & \best{48.10} & 80.59 \\
    HMoDE~\cite{hmode} & TIP'23 & 57.36 & 80.89 \\
    HMoDE+REL~\cite{hmode} & TIP'23 & 50.91 & \second{76.64} \\
    \midrule
    \textbf{Count Anything} & -- & 55.90 & 92.15 \\
    \bottomrule
    \end{tabular*}
    \vspace{-6pt}
\end{table}

\begin{table}[t]
    \centering
    \scriptsize
    \setlength{\tabcolsep}{1.4pt}
    \renewcommand{\arraystretch}{0.8}
    \captionsetup{position=top,skip=1pt}
    \caption{Cross-dataset transfer from FSC-147 to CLOC. All models trained on FSC-147 are evaluated on the CLOC test set.}
    \label{tab:comp_fsc147}
    \begin{tabular*}{\columnwidth}{@{\extracolsep{\fill}}llcc@{}}
    \toprule
    \textbf{Method} & \textbf{Venue} & \textbf{MAE} & \textbf{RMSE} \\
    \midrule
    LOCA~\cite{loca} & ICCV'23 & 68.86 & 179.26 \\
    CLIP-Count~\cite{clip_count} & ACMMM'23 & 31.28 & \second{118.55} \\
    CounTX~\cite{countx} & BMVC'23 & 32.24 & 122.10 \\
    VLCounter~\cite{vlcounter} & AAAI'24 & 30.52 & 121.88 \\
    VA-Count~\cite{va_count} & ECCV'24 & 33.85 & 129.81 \\
    CountGD~\cite{countgd} & NeurIPS'24 & 55.91 & 1081.82 \\
    YOLO-Count~\cite{yolo_count} & ICCV'25 & 34.15 & 133.56 \\
    T2ICount~\cite{t2icount} & CVPR'25 & 50.02 & 155.92 \\
    CountSE~\cite{countse} & ICCV'25 & 28.81 & 130.49 \\
    CountGD++~\cite{countgdpp} & CVPR'26 & \second{23.38} & 123.28 \\
    \midrule
    \textbf{Count Anything} & -- & \best{19.01} & \best{100.12} \\
    \bottomrule
    \end{tabular*}
    \vspace{-14pt}
\end{table}

\textbf{Dense crowd counting performance.}
Crowd counting is a widely studied branch of object counting, and it is also one of the most challenging subtasks due to extremely dense target distributions and complex occlusion relationships. Therefore, we conduct an additional evaluation on the dense crowd counting subset from ShanghaiTech (SHTech) Part A in the CLOC test split. Table~\ref{tab:comp_crowd} compares Count Anything with existing open-world counting models and typical crowd-counting-specific methods. As can be seen, compared with open-world baselines, Count Anything substantially reduces MAE and RMSE. Compared with methods specifically designed for crowd counting, Count Anything also achieves performance within the same tier, although it slightly lags behind the most advanced crowd-specific methods. Considering that these crowd-specific methods are usually specially optimized for a single crowd dataset and high-density distributions, while Count Anything is designed for a unified text-guided counting task across categories and visual domains, this result shows that Count Anything can achieve strong competitiveness on the crowd counting subset while maintaining generality.

\begin{table}[t]
    \centering
    \scriptsize
    \setlength{\tabcolsep}{3pt}
    \renewcommand{\arraystretch}{0.8}
    \captionsetup{position=top,skip=1pt}
    \caption{Comparison under shared CLOC training data. The Seen/Unseen refer to semantic category groups included in/excluded from CLOC training.}
    \label{tab:comp_cloc}
    \begin{tabular*}{\columnwidth}{@{\extracolsep{\fill}}lccc@{}}
        \toprule
        \multirow{2}{*}{\textbf{Method}} &
        \multicolumn{3}{c}{\textbf{Full Test MAE}} \\
        \cmidrule(lr){2-4}
        & \textbf{Overall} & \textbf{Seen} & \textbf{Unseen} \\
        \midrule
        CLIP-Count~\cite{clip_count} & 14.48 & 19.16 & 7.24 \\
        VLCounter~\cite{vlcounter} & 21.61 & 28.68 & 10.68 \\
        CountGD++~\cite{countgdpp} & 14.71 & 20.64 & 5.56 \\
        \midrule
        \textbf{Count Anything} & \best{9.34} & \best{12.37} & \best{4.79} \\
        \bottomrule
    \end{tabular*}
    \vspace{-7pt}
\end{table}

\textbf{Comparison under aligned training data.}
To account for training-data differences, we compare two aligned settings. Table~\ref{tab:comp_fsc147} evaluates FSC-147-trained methods on CLOC. Count Anything uses the official FSC-147 training split; other methods use their officially released FSC-147-trained weights. Notably, CLOC preserves FSC-147's official split: training images enter only CLOC training, and test images only CLOC testing. This keeps FSC-147 training and test images disjoint while all methods are evaluated on the full CLOC test set. Under this setting, Count Anything achieves 19.01 MAE and 100.12 RMSE, outperforming representative FSC-147-trained counting methods. Second, in Table~\ref{tab:comp_cloc}, we place representative methods under the same CLOC training setting, train them on the full CLOC training split using their official configurations, and evaluate them under the same full-test protocol. The seen/unseen split of CLOC is based on semantic category groups, where unseen categories are excluded from training to evaluate category generalization; details are provided in Appendix~\ref{app:task_aware_splitting}. Under this setting, Count Anything obtains the lowest overall, seen, and unseen MAE. Together, these two comparisons show that the performance advantage of Count Anything is not merely due to training-data differences, but also reflects the effectiveness of the proposed model design for cross-domain text-guided counting.

\begin{table}[t]
    \centering
    \scriptsize
    \setlength{\tabcolsep}{2pt}
    \renewcommand{\arraystretch}{0.8}
    \captionsetup{position=top,skip=1pt}
    \caption{Ablation of counting branches and fusion on CLOC. Here, DU denotes direct union of RSC and PDC predictions.}
    \label{tab:ablation_dual_counter_fusion}
    \begin{tabular*}{\columnwidth}{@{\extracolsep{\fill}}cccccc@{}}
        \toprule
        \multicolumn{2}{c}{\textbf{Counter}} &
        \multicolumn{2}{c}{\textbf{Fusion}} &
        \multicolumn{2}{c}{\textbf{Metric}} \\
        \cmidrule(lr){1-2}
        \cmidrule(lr){3-4}
        \cmidrule(lr){5-6}
        \textbf{RSC} & \textbf{PDC} &
        \textbf{DU} & \textbf{CCF} &
        \textbf{MAE} & \textbf{RMSE} \\
        \midrule
        \cmark & \xmark & \xmark & \xmark & 18.66 & 116.42 \\
        \xmark & \cmark & \xmark & \xmark & 13.30 & 35.31 \\
        \cmark & \cmark & \cmark & \xmark & 22.85 & 50.05 \\
        \cmark & \cmark & \xmark & \cmark & \best{9.34} & \best{33.34} \\
        \bottomrule
    \end{tabular*}
    \vspace{-7pt}
\end{table}

\begin{table}[t]
    \centering
    \scriptsize
    \setlength{\tabcolsep}{2.2pt}
    \renewcommand{\arraystretch}{0.8}
    \captionsetup{position=top,skip=1pt}
    \caption{Point-centric supervision ablations on CLOC.}
    \label{tab:ablation_supervision_strategies}
    \begin{tabular*}{\columnwidth}{@{\extracolsep{\fill}}cccccc@{}}
        \toprule
        \multicolumn{2}{c}{\textbf{Box/GIoU sup.}} &
        \multicolumn{2}{c}{\textbf{RSC cls. sup.}} &
        \multicolumn{2}{c}{\textbf{Metric}} \\
        \cmidrule(lr){1-2}
        \cmidrule(lr){3-4}
        \cmidrule(lr){5-6}
        \textbf{GT} & \textbf{Pseudo} &
        \textbf{Hard} & \textbf{Soft} &
        \textbf{MAE} & \textbf{RMSE} \\
        \midrule
        \cmark & \cmark & \xmark & \cmark & 11.44 & 38.20 \\
        \cmark & \xmark & \cmark & \xmark & 17.36 & 92.70 \\
        \cmark & \xmark & \xmark & \cmark & \best{9.34} & \best{33.34} \\
        \bottomrule
    \end{tabular*}
    \vspace{-14pt}
\end{table}

\vspace{-4pt}
\subsection{Ablation Study}
\label{sec:ablation_studies}
\vspace{-4pt}

\textbf{Dual counters and fusion rules.}
Table~\ref{tab:ablation_dual_counter_fusion} compares the two branches and fusion rules; DU denotes direct union. RSC-only follows SAM3's original region-prediction paradigm and yields 18.66 MAE, well above Count Anything's 9.34. This indicates that task-adapted region-level prediction alone remains insufficient for accurate cross-domain counting, supporting PDC-based dense instance enumeration and the branches' complementarity across object scales and densities. PDC-only also underperforms the full model. Direct union causes duplicate counts and larger errors, while CCF achieves the best performance, validating the dual-counter design and complementary fusion. Further ablations on the number of PDC candidates compared in each RSC region are provided in Appendix~\ref{app:additional_results} (Table~\ref{tab:ablation_ccf_budget}).

\textbf{Point-centric supervision strategies.}
Table~\ref{tab:ablation_supervision_strategies} ablates two design choices in point-centric supervision: the box targets used for RSC Box/GIoU supervision and the foreground targets used for RSC classification. Following Sec.~\ref{sec:method}, Count Anything uses GT-only box targets for geometric regression and a soft classification target for matched RSC queries. Adding pseudo box targets to point-only samples increases error, indicating that pseudo boxes may introduce noise that degrades region-level geometry learning. Replacing the soft classification target with a hard foreground target also leads to worse performance, showing that quality-aware foreground supervision helps RSC distinguish high-quality region candidates from lower-quality ones. These results support the point-centric supervision design: all instances contribute through point supervision, reliable GT boxes supervise geometry, and soft foreground targets guide RSC confidence learning.

\begin{figure}[t]
    \centering
    \includegraphics[width=0.8\linewidth]{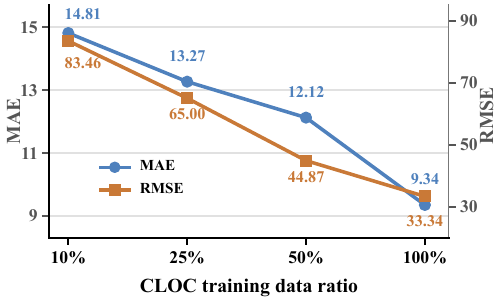}
    \vspace{-12pt}
\caption{Effect of the training data scale on Count Anything.}    \label{fig:cloc_scaling}
    \vspace{-14pt}
\end{figure}

\textbf{Data scaling.}
Figure~\ref{fig:cloc_scaling} visualizes the effect of CLOC training data scale by plotting MAE and RMSE under different ratios. As the data ratio increases from 10\% to 100\%, both MAE and RMSE consistently decrease, showing that Count Anything benefits from larger-scale cross-domain counting data. This result supports the value of data scaling-up: the full CLOC training set provides broader category, visual-domain, and density coverage, leading to a more stable training basis for generalist text-guided counting.

\vspace{-4pt}
\section{Conclusion}
\vspace{-4pt}
This paper formulates generalist object counting as instance-grounded enumeration of user-specified targets across visual domains.
We construct CLOC, a large-scale dataset covering six visual domains, with a unified counting representation for heterogeneous annotations and broad count and density distributions.
We propose Count Anything, a text-guided model that combines region-level anchoring with dense point localization through dual-granularity instance enumeration.
Point-centric supervision enables joint learning from point and box annotations, while Complementary Count Fusion integrates the branch predictions into a unified point set.
Experiments on CLOC demonstrate leading counting accuracy across all six domains and strong instance localization.

{
    \small
    \bibliographystyle{ieeenat_fullname}
    \bibliography{a_cite}
}

\clearpage
\onecolumn
\appendix
\addtocontents{toc}{\protect\setcounter{tocdepth}{3}}
\maketitleappendix
\setcounter{tocdepth}{3}
\tableofcontents
\setcounter{tocdepth}{3}
\twocolumn
\raggedbottom
\setcounter{secnumdepth}{3}
\emergencystretch=1em
\microtypesetup{activate=true}

\setcounter{figure}{0}
\setcounter{table}{0}
\setcounter{equation}{0}
\renewcommand{\thefigure}{A\arabic{figure}}
\renewcommand{\thetable}{A\arabic{table}}
\renewcommand{\theequation}{A\arabic{equation}}
\renewcommand{\theHfigure}{app.\arabic{figure}}
\renewcommand{\theHtable}{app.\arabic{table}}
\renewcommand{\theHequation}{app.\arabic{equation}}

\setcounter{lstlisting}{0}
\renewcommand{\lstlistingname}{Code}
\renewcommand{\thelstlisting}{A\arabic{lstlisting}}
\providecommand{\theHlstlisting}{app.\arabic{lstlisting}}
\renewcommand{\theHlstlisting}{app.\arabic{lstlisting}}

\section{Additional Experimental Settings}
\label{app:implementation_details}

\subsection{Implementation Details}

The text-conditioned encoder of Count Anything is initialized from the pretrained SAM3 model. During training, the original parameters of the pretrained text-conditioned encoder are frozen. We insert trainable LoRA adapters into the cross-modal encoder for lightweight adaptation, with rank $r=8$, scaling factor $\alpha=8$. The Region-level Sparse Counter (RSC) follows the region-query decoding formulation in Sec.~\ref{sec:method} and uses $Q_r=200$ region queries. Each region query predicts a foreground logit and a candidate box, whose center is used as the RSC counting point.
For the Pixel-level Dense Counter (PDC), we use the dense spatial grid $\Omega=\{u_k\}_{k=1}^{Q_d}$ and anchor point $a_k=\pi(u_k)$ defined in Sec.~\ref{sec:method}. In our implementation, each grid location generates one PDC candidate, and adjacent anchor points are spaced by $7$ pixels in the input-image coordinate system. The PDC regression head predicts a two-dimensional offset $\Delta_k$ relative to the anchor point. The offset scaling factor is set to $\rho=100$, so the predicted PDC point is computed as
\begin{equation}
    \hat{p}_k^d = a_k + 100\Delta_k .
\end{equation}
The PDC feature adapter contains $3$ residual blocks with $256$ channels and group normalization.

Training is performed with distributed data parallelism using the NCCL backend. The per-GPU training batch size is $24$, and the gradient accumulation step is $1$, resulting in an effective training batch size of $192$. We train for $30$ epochs and evaluate after each epoch. The random seed is fixed to $123$. The optimizer is AdamW with $(\beta_1,\beta_2)=(0.9,0.999)$, $\epsilon=10^{-8}$, and weight decay $10^{-4}$. We use \texttt{bfloat16} automatic mixed precision. Gradients are clipped by the global $\ell_2$ norm with a maximum norm of $0.1$.

The learning rate is scheduled by an epoch-wise cosine scheduler from epoch $0$ to epoch $30$ without warm-up. The minimum learning rate ratio is $0.1$. The initial learning rates are $1\times10^{-5}$ for RSC, $1\times10^{-4}$ for PDC, and $1\times10^{-3}$ for LoRA adapters. Accordingly, the final learning rates after cosine decay are $1\times10^{-6}$, $1\times10^{-5}$, and $1\times10^{-4}$ for the three parameter groups, respectively.

All samples are trained and evaluated at an input resolution of $1008\times1008$. During training, we randomly choose one of two preprocessing paths with equal probability. The first path applies aspect-ratio-preserving random resizing with a scale sampled from $[0.5,2.0]$, constrains the shorter side to be at least $256$ pixels and the longer side to be at most $1500$ pixels, and then randomly crops or pads the image to $1008\times1008$. The second path directly resizes the image to a square input of $1008\times1008$ and pads it when needed. Validation and testing use deterministic resize/pad preprocessing. Images are converted to tensors and normalized with mean $(0.5,0.5,0.5)$ and standard deviation $(0.5,0.5,0.5)$. Category names are used as text queries, and each image loads at most $8$ training or validation queries.

For branch-specific matching, RSC uses Hungarian matching. Its matching cost consists of a foreground confidence cost and a point-distance cost, whose weights are $1.0$ and $0.05$, respectively. PDC uses sparse Hungarian matching. For each ground-truth point, the local candidate set $\mathcal{N}(j)$ contains the top-$5$ nearby dense candidates. The PDC matching cost also uses foreground confidence and point distance, with weights $1.0$ and $0.05$, respectively.

The overall loss follows the point-centric supervision in Sec.~\ref{sec:method}. For RSC, the weights of classification loss, point loss, box loss, and GIoU loss are $20.0$, $5.0$, $5.0$, and $2.0$, respectively. The soft foreground target in RSC classification uses quality-balancing coefficient $\alpha=0.25$ and minimum positive label value $\epsilon=0.01$. For PDC, the weights of classification loss and point-regression loss are $1.0$ and $2\times10^{-4}$, respectively.

During inference, CCF uses score thresholds $\tau_r=0.5$ and $\tau_d=0.5$ for RSC and PDC candidates. RSC duplicate removal processes RSC boxes in descending order of score. A lower-scored region is removed if it is fully contained by a retained region or if its intersection over minimum area (IoM) with a retained region is larger than $0.5$. After RSC filtering, CCF first identifies, for each retained RSC box, the PDC point inside the box that is closest to the corresponding RSC center. It then compares the RSC score with the score of this nearest PDC point and suppresses the lower-scored one. In this way, CCF compares each retained RSC box with at most one PDC point based on their confidence scores, while preserving the other PDC points inside the box during this comparison. The remaining RSC and PDC points are merged as the final prediction, and the final count is the size of the fused point set. 

\subsection{Evaluation Protocol}
\label{app:evaluation_protocol}

CLOC is primarily designed for text-guided object counting evaluation. In practice, category names are used as textual queries, so each image-category pair corresponds to an image-query counting sample. Given an image and a target query, a model is required to predict the number of instances corresponding to that query in the image. For evaluation, we use MAE and RMSE to measure counting errors and density-normalized average precision (nAP)~\cite{p2pnet} to evaluate instance localization accuracy.

Assume the test set contains $N$ image-category samples. For the $i$-th sample, let $c_i$ denote the ground-truth count and $\hat{c}_i$ denote the predicted count.

MAE measures the mean absolute counting error:
\[
\mathrm{MAE}=\frac{1}{N}\sum_{i=1}^{N}|c_i-\hat{c}_i|.
\]
MAE directly reflects the average number of targets incorrectly predicted per sample, and is the most intuitive counting-error metric.

RMSE measures the root mean squared error:
\[
\mathrm{RMSE}=\sqrt{\frac{1}{N}\sum_{i=1}^{N}(c_i-\hat{c}_i)^2}.
\]
RMSE is more sensitive to large errors, and therefore reflects model stability on high-density scenes or difficult samples. When a model produces severe counting errors on a small number of samples, RMSE highlights such failures more clearly than MAE.

nAP measures instance localization accuracy. Following P2PNet~\cite{p2pnet}, we normalize the localization error between a predicted point $\hat p_i$ and a ground-truth point $p_j$ by $r_j$, the mean distance from $p_j$ to its three nearest other ground-truth points:
\[
d_{ij}=\frac{\|\hat p_i-p_j\|_2}{r_j}.
\]
Within each image-category query, predictions are matched one-to-one in descending order of score, and a match is considered correct if $d_{ij}<\delta$. For localization evaluation, we exclude nonempty queries with fewer than four ground-truth points or invalid local scales. Empty queries are retained, and all predictions for these queries are counted as false positives. All methods are evaluated on the same 7,720 valid image-category queries, whose predictions are pooled to compute AP. Counting metrics are computed on the full test set. We report $\mathrm{nAP}_{25}$ and $\mathrm{nAP}_{50}$ at thresholds of 0.25 and 0.50, respectively, and $\mathrm{nAP}_{05\,:\,50}$ averaged over thresholds from 0.05 to 0.50 in steps of 0.05. Higher values indicate better localization performance.

The basic evaluation unit of CLOC is an image-category pair, rather than the total number of targets in an entire image. If an image contains multiple target categories, each category forms an independent counting sample. This setting matches text-guided category counting, where the model needs to output the number of instances for the specified query rather than estimate the total number of all visible targets in the image.

\subsection{Experimental Fairness Analysis}
\label{sec:supp_fairness}

\paragraph{Out-of-the-box evaluation.}
The competing methods in Table~\ref{tab:merged_full_domain_mae} are released as open-world, open-vocabulary, or
generalist models, which are intended to be applied to new categories and scenes
without further training. We therefore evaluate them directly with their official
checkpoints and default inference settings, reflecting how these models are expected
to be used in practice. All methods are evaluated on the same image--category queries
of the CLOC test split, using the same category names as text queries and the same
ground-truth annotations.

\paragraph{Aligned training data.}
Since Count Anything is trained on CLOC, Table~\ref{tab:merged_full_domain_mae} reflects differences in both model
design and training data. The main paper disentangles these two factors through two
comparisons with aligned training data. In Table~\ref{tab:comp_fsc147}, Count Anything is trained only on
the official FSC-147 training split and compared with counting methods trained on
FSC-147. In Table~\ref{tab:comp_cloc}, the competing methods are retrained on the same CLOC training
split using their official configurations. Count Anything remains the best in both
settings, indicating that its advantage does not stem solely from large-scale training
data; its dual-granularity enumeration and the corresponding supervision strategy are also essential to its strong performance (Tables~\ref{tab:ablation_dual_counter_fusion} and~\ref{tab:ablation_supervision_strategies}).

\paragraph{Split integrity.}
The derived cropped and stitched samples are generated only from training images and
never enter the validation or test splits. For video-derived datasets, frames are
temporally subsampled to reduce near-duplicates. The official FSC-147 split is also
preserved, so that its training and test images remain disjoint in CLOC. Any overlap
between the pretraining data of competing models and the CLOC test split could only
benefit these models, and therefore does not inflate the advantage of Count Anything.

\begin{table*}[t]
\centering
\scriptsize
\setlength{\tabcolsep}{1.2pt}
\renewcommand{\arraystretch}{1.15}
\captionsetup{position=top,skip=3pt}
\caption{Extended comparison with other methods on
CLOC. The first block lists object counting methods, and the second
lists general-purpose detection, segmentation, and visual grounding
models. Best and second-best results are boldfaced and underlined,
respectively.}
\label{tab:app_merged_full_domain_mae}

\begin{tabular*}{\textwidth}{@{\extracolsep{\fill}}llc*{7}{rr}@{}}
\toprule
\multirow{2}{*}{\textbf{Method}} &
\multirow{2}{*}{\textbf{Venue}} &
\multirow{2}{*}{\makecell{\textbf{Ins.}\\\textbf{Grd.}}} &
\multicolumn{2}{c}{\textbf{Full Test}} &
\multicolumn{2}{c}{\textbf{GS}} &
\multicolumn{2}{c}{\textbf{RS}} &
\multicolumn{2}{c}{\textbf{Histo.}} &
\multicolumn{2}{c}{\textbf{Cell.}} &
\multicolumn{2}{c}{\textbf{Agri.}} &
\multicolumn{2}{c}{\textbf{Micro.}} \\
\cmidrule(lr){4-5}
\cmidrule(lr){6-7}
\cmidrule(lr){8-9}
\cmidrule(lr){10-11}
\cmidrule(lr){12-13}
\cmidrule(lr){14-15}
\cmidrule(lr){16-17}
& & &
MAE & RMSE &
MAE & RMSE &
MAE & RMSE &
MAE & RMSE &
MAE & RMSE &
MAE & RMSE &
MAE & RMSE \\
\midrule

LOCA~\cite{loca} & ICCV'23 & \xmark &
68.86 & 179.26 &
55.24 & 160.90 & 81.37 & 168.77 &
65.86 & 120.86 & 124.36 & 261.68 &
473.81 & 620.06 & 39.47 & 63.74 \\

CLIP-Count~\cite{clip_count} & ACMMM'23 & \xmark &
31.28 & \second{118.55} &
24.07 & 71.65 & 25.81 & 44.40 &
57.98 & 94.05 & 121.17 & 242.74 &
298.45 & 748.52 & 22.40 & 37.29 \\

CounTX~\cite{countx} & BMVC'23 & \xmark &
32.24 & 122.10 &
23.84 & 70.47 & 30.06 & 60.20 &
64.64 & 121.88 & \second{112.22} & \second{239.13} &
301.10 & 767.27 & 27.41 & 41.62 \\

VLCounter~\cite{vlcounter} & AAAI'24 & \xmark &
30.52 & 121.88 &
23.59 & 71.77 & 22.45 & 38.57 &
55.14 & 94.05 & 125.83 & 247.36 &
306.38 & 783.42 & 37.72 & 53.48 \\

VA-Count~\cite{va_count} & ECCV'24 & \xmark &
33.85 & 129.81 &
22.39 & 73.66 & 36.62 & 76.11 &
72.50 & 119.17 & 163.73 & 366.38 &
315.23 & 764.95 & 20.47 & 38.47 \\

CountGD~\cite{countgd} & NeurIPS'24 & \cmark &
55.91 & 1081.82 &
53.83 & 1288.10 & 42.23 & 74.48 &
99.69 & 154.05 & 114.50 & 260.35 &
\second{244.80} & \second{604.77} & 11.09 & 22.80 \\

YOLO-Count~\cite{yolo_count} & ICCV'25 & \xmark &
34.15 & 133.56 &
25.62 & 89.31 & 27.69 & 63.21 &
72.50 & 122.79 & 168.37 & 293.50 &
309.30 & 778.96 & 24.33 & 35.43 \\

T2ICount~\cite{t2icount} & CVPR'25 & \xmark &
50.02 & 155.92 &
34.08 & 117.59 & 62.46 & 116.48 &
73.06 & 142.75 & 151.30 & 266.74 &
296.37 & 749.84 & 203.51 & 265.96 \\

CountSE~\cite{countse} & ICCV'25 & \xmark &
28.81 & 130.49 &
22.14 & 87.65 & 19.07 & 41.13 &
70.27 & 121.15 & 120.08 & 249.88 &
302.19 & 790.75 & 22.22 & 40.20 \\

CountGD++~\cite{countgdpp} & CVPR'26 & \cmark &
23.38 & 123.28 &
\second{14.50} & \second{56.98} &
10.69 & \second{22.92} &
55.83 & 115.62 & 222.25 & 397.72 &
321.99 & 792.34 & 43.80 & 61.96 \\

SAM3Count~\cite{sam3count} & CVPRW'26 & \cmark &
22.84 & 122.64 &
16.20 & 66.58 & 10.25 & 32.61 &
63.11 & 125.88 & 128.50 & 292.41 &
307.30 & 791.24 & 34.44 & 55.89 \\

\midrule

GLIP-L~\cite{glip} & CVPR'22 & \cmark &
37.38 & 140.07 &
31.06 & 94.67 & 19.57 & 44.35 &
68.78 & 129.88 & 225.42 & 398.12 &
320.85 & 791.92 & 43.71 & 61.63 \\

YOLO-World-X~\cite{yolo_world} & CVPR'24 & \cmark &
34.94 & 138.35 &
28.17 & 91.48 & 17.81 & 41.78 &
67.93 & 127.93 & 226.07 & 398.14 &
321.13 & 791.97 & 42.49 & 61.05 \\

GrdDINO-SwinB~\cite{grounding_dino} & ECCV'24 & \cmark &
35.06 & 139.46 &
29.50 & 94.38 & 14.13 & 37.53 &
69.11 & 130.12 & 224.96 & 397.40 &
320.17 & 791.83 & 43.75 & 61.61 \\

YOLOE-v8L~\cite{yoloe} & ICCV'25 & \cmark &
40.03 & 140.62 &
33.97 & 95.46 & 21.78 & 45.66 &
69.99 & 130.60 & 228.28 & 399.19 &
321.97 & 791.94 & 44.80 & 62.39 \\

YOLOE-11L~\cite{yoloe} & ICCV'25 & \cmark &
40.05 & 140.60 &
34.00 & 95.47 & 21.78 & 45.66 &
69.95 & 130.62 & 228.24 & 399.17 &
321.88 & 791.70 & 44.81 & 62.40 \\

SAM3~\cite{sam3} & ICLR'26 & \cmark &
25.80 & 130.26 &
19.39 & 79.86 & 10.88 & 29.08 &
64.32 & 125.79 & 152.45 & 351.54 &
315.88 & 792.02 & 37.04 & 53.69 \\

GroundAnything-4B~\cite{groundanything} & arXiv'26 & \cmark &
25.67 & 130.56 &
19.51 & 81.37 & 11.56 & 28.19 &
63.24 & 118.09 & 163.01 & 354.58 &
299.05 & 790.64 & 19.78 & 34.05 \\

WeVisOne-VL-2B~\cite{wevisone} & arXiv'26 & \cmark &
\second{22.51} & 128.14 &
17.52 & 79.41 & \second{7.88} & 24.13 &
\second{43.72} & \second{90.65} & 163.18 & 350.91 &
287.70 & 785.06 & \second{7.16} & \second{20.17} \\

MolmoPoint-8B~\cite{molmopoint} & ECCV'26 & \cmark &
29.29 & 136.64 &
23.49 & 89.87 & 11.46 & 35.91 &
63.06 & 124.25 & 217.51 & 395.27 &
300.01 & 788.93 & 19.56 & 44.27 \\

LocateAnything-3B~\cite{locateanything} & ECCV'26 & \cmark &
31.77 & 141.52 &
26.27 & 96.62 & 12.35 & 37.59 &
76.00 & 176.22 & 198.35 & 388.59 &
305.21 & 791.64 & 31.18 & 49.01 \\

\midrule

\textbf{Count Anything} & -- & \cmark &
\best{9.34} & \best{33.34} &
\best{8.76} & \best{29.96} &
\best{6.21} & \best{21.07} &
\best{16.64} & \best{33.61} &
\best{38.65} & \best{92.75} &
\best{41.78} & \best{117.95} &
\best{4.30} & \best{12.89} \\

\bottomrule
\end{tabular*}

\par\vspace{3pt}
{\scriptsize\raggedright
\noindent\textit{Note:} Ins.Grd. denotes instance grounding capability.
GS, RS, Histo., Cell., Agri., and Micro. denote General Scene,
Remote Sensing, Histopathology, Cellular Microscopy, Agriculture,
and Microbiology, respectively.
Lower MAE and RMSE indicate better performance.
\par}
\vspace{-10pt}
\end{table*}

\begin{table}[t]
    \centering
    \small
    \setlength{\tabcolsep}{2.2pt}
    \renewcommand{\arraystretch}{1.15}
    \captionsetup{position=top,skip=1pt}
    \caption{Extended comparison of instance localization performance
    on CLOC. The first block lists general-purpose detection,
    segmentation, and visual grounding models, and the second
    lists generalist counting methods. Best and second-best
    results are boldfaced and underlined, respectively.}
    \label{tab:supp_comp_localization}
    \begin{tabular*}{\columnwidth}{@{\extracolsep{\fill}}lccc@{}}
        \toprule
        \textbf{Method} &
        $\mathbf{nAP}_{25}$ &
        $\mathbf{nAP}_{50}$ &
        $\mathbf{nAP}_{05\,: \,50}$ \\
        \midrule
        GrdDINO-SwinB~\cite{grounding_dino}
        & 5.34 & 6.81 & 5.30 \\
        YOLOE-v8L~\cite{yoloe}
        & 2.70 & 2.84 & 2.59 \\
        YOLOE-11L~\cite{yoloe}
        & 2.75 & 3.77 & 2.81 \\
        SAM3~\cite{sam3}
        & 23.63 & 26.42 & 21.34 \\
        GroundAnything~\cite{groundanything}
        & 21.67 & 28.46 & 20.24 \\
        LocateAnything-3B~\cite{locateanything}
        & 12.15 & 15.75 & 11.31 \\
        \midrule
        CountGD~\cite{countgd}
        & 30.35 & 40.55 & 27.90 \\
        CountGD++~\cite{countgdpp}
        & 35.86 & 45.43 & 32.09 \\
        SAM3Count~\cite{sam3count}
        & \underline{38.28} & \underline{46.21} & \underline{34.64} \\
        \midrule
        \textbf{Count Anything}
        & \textbf{53.93} & \textbf{70.27} & \textbf{50.26} \\
        \bottomrule
    \end{tabular*}
    \par\vspace{5pt}
    \vspace{-10pt}
\end{table}

\begin{table}[t]
    \centering
    \small
    \setlength{\tabcolsep}{3pt}
    \renewcommand{\arraystretch}{1.15}
    \captionsetup{position=top,skip=1pt}
    \caption{Extended comparison under shared CLOC training data. Seen and Unseen denote semantic category groups included in and excluded from CLOC training, respectively.}
    \label{tab:supp_comp_cloc}

    \begin{tabular*}{\columnwidth}{@{\extracolsep{\fill}}lccc@{}}
        \toprule
        \multirow{2}{*}{\textbf{Method}} &
        \multicolumn{3}{c}{\textbf{Full Test MAE}} \\
        \cmidrule(lr){2-4}
        & \textbf{Overall} & \textbf{Seen} & \textbf{Unseen} \\
        \midrule
        CLIP-Count~\cite{clip_count}
        & 14.48 & 19.16 & 7.24 \\
        VLCounter~\cite{vlcounter}
        & 21.61 & 28.68 & 10.68 \\
        CountGD++~\cite{countgdpp}
        & 14.71 & 20.64 & 5.56 \\
        SAM3~\cite{sam3}
    & 24.65 & 35.74 & 7.53 \\
       LocateAnything-3B~\cite{locateanything}
        & 21.36 & 31.96 & 5.00 \\
        \midrule
        \textbf{Count Anything}
        & \best{9.34} & \best{12.37} & \best{4.79} \\
        \bottomrule
    \end{tabular*}
\end{table}

\section{Additional Experimental Results}
\label{app:additional_results}
\subsection{Quantitative Experiments}

\paragraph{Extended main results on CLOC.}
Table~\ref{tab:app_merged_full_domain_mae} supplements Table~\ref{tab:merged_full_domain_mae} with domain-wise RMSE results.
Our Count Anything achieves the lowest MAE and RMSE on the full test set and across all six visual domains, with particularly substantial improvements in histopathology, cellular microscopy, and agriculture.
Its overall MAE of 9.34 and RMSE of 33.34 demonstrate strong counting accuracy and reduced large-error predictions across diverse counting scenarios.
For CountGD, the pronounced MAE-RMSE gap is consistent with a small number of large counting errors whose influence is amplified by the squared-error aggregation in RMSE.

\paragraph{Extended instance localization comparison.}
Table~\ref{tab:supp_comp_localization} extends the instance localization comparison to a broader range of general-purpose detection, segmentation, and visual grounding models, as well as generalist counting methods.
All methods are evaluated using the same localization protocol described in Sec.~\ref{app:evaluation_protocol}.
Count Anything achieves the best results across all three metrics, obtaining 53.93, 70.27, and 50.26 on $\mathrm{nAP}_{25}$, $\mathrm{nAP}_{50}$, and $\mathrm{nAP}_{05\,: \,50}$, respectively.
Compared with the strongest competing method, SAM3Count, it improves $\mathrm{nAP}_{05\,: \,50}$ by 15.62 points.
These results complement the counting evaluation, showing that Count Anything achieves accurate instance localization alongside low counting errors on CLOC.

\paragraph{Extended comparison under shared CLOC training data.}
For CLIP-Count, VLCounter, and CountGD++, we retrain the models on the CLOC training split following their official initialization and training procedures.
For SAM3, we perform LoRA fine-tuning on CLOC. Its MAE remains higher than that of Count Anything, further supporting the effectiveness of our method.
For LocateAnything-3B, we follow its official LoRA fine-tuning configuration and train the LoRA parameters of the language model and the MLP projector on CLOC.
Under these CLOC-based training settings, Count Anything achieves the lowest overall, seen, and unseen MAE (Table~\ref{tab:supp_comp_cloc}).

\begin{table}[!t]
\centering
\small
\setlength{\tabcolsep}{3pt}
\renewcommand{\arraystretch}{1.15}
\caption{Model complexity and inference efficiency of Count Anything under 1008$\times$1008 input resolution. Latency and memory are measured with batch size 1 using BF16 inference.}
\label{tab:complexity_latency}

\begin{tabular*}{\columnwidth}{@{\extracolsep{\fill}}lcccrr@{}}
\toprule
\textbf{Model} & \textbf{Params} & \textbf{FLOPs} & \textbf{GPU} &
\makecell[r]{\textbf{Latency}\\\textbf{(ms)}} &
\makecell[r]{\textbf{Memory}\\\textbf{(GB)}} \\
\midrule
 & & & RTX 5090 & 68.51 & 5.17 \\
 & & & RTX 4090 & 95.55 & 6.53 \\
\multirow{-3}{*}{\makecell[l]{Count\\Anything}} &
\multirow{-3}{*}{848.62M} &
\multirow{-3}{*}{5.44T} &
A100 & 122.11 & 6.53 \\
\bottomrule
\end{tabular*}
\end{table}

\paragraph{Computational efficiency.}
We evaluate the computational efficiency of Count Anything under the input resolution of 1008$\times$1008. As shown in Table~\ref{tab:complexity_latency}, Count Anything contains 848.62M parameters and requires 5.44 TFLOPs per forward pass. Despite operating on high-resolution images with dual counting branches, the model maintains practical inference efficiency on modern GPUs. Specifically, it achieves an average latency of 68.51 ms on an RTX 5090, 95.55 ms on an RTX 4090, and 122.11 ms on an A100, with peak inference memory ranging from 5.17 GB to 6.53 GB. These results indicate that Count Anything can provide high-resolution text-guided counting while maintaining feasible computational cost for deployment.

\begin{table}[!t]
\centering
\small
\setlength{\tabcolsep}{3pt}
\renewcommand{\arraystretch}{1.15}

\caption{Ablation on point-centric supervision components.}
\label{tab:ablation_supervision_components}
\begin{tabular*}{\columnwidth}{@{\extracolsep{\fill}}ccrr@{}}
\toprule
\multicolumn{2}{c}{\textbf{Supervision}} & \multicolumn{2}{c}{\textbf{Metric}} \\
\cmidrule(lr){1-2}
\cmidrule(lr){3-4}
\textbf{RSC point} & \textbf{Box and GIoU} & \textbf{MAE} & \textbf{RMSE} \\
\midrule
\xmark & \cmark & 14.06 & 89.80 \\
\cmark & \xmark & 13.83 & 61.92 \\
\cmark & \cmark & \textbf{9.34} & \textbf{33.34} \\
\bottomrule
\end{tabular*}
\end{table}
\paragraph{Point-centric supervision components.}
Table~\ref{tab:ablation_supervision_components} studies the effect of the core supervision components in point-centric supervision. Removing the RSC point loss weakens RSC's ability to learn counting-point localization from point-only annotations, leading to substantially larger errors. Removing the box and GIoU losses weakens the region-level geometric supervision of RSC, especially for domains where object extent provides useful anchoring. Using both RSC point supervision and box-level geometric supervision achieves the best performance, showing that point-level localization and box-level geometry provide complementary supervision for RSC.

\begin{table}[!t]
\centering
\small
\setlength{\tabcolsep}{3pt}
\renewcommand{\arraystretch}{1.15}

\caption{Ablation on the PDC feature adapter.}
\label{tab:ablation_pdc_adapter}

\begin{tabular*}{\columnwidth}{@{\extracolsep{\fill}}crr@{}}
\toprule
\multirow{2}{*}{\textbf{Adapter design}} & \multicolumn{2}{c}{\textbf{Metric}} \\
\cmidrule(lr){2-3}
& \textbf{MAE} & \textbf{RMSE} \\
\midrule
None & 12.48 & 48.25 \\
1 residual block & 11.04 & 41.20 \\
\textbf{3 residual blocks} & \textbf{9.34} & \textbf{33.34} \\
\bottomrule
\end{tabular*}
\end{table}

\paragraph{PDC feature adapter.}
Table~\ref{tab:ablation_pdc_adapter} studies the effect of the feature adapter in PDC. Without the adapter, PDC directly performs point prediction from the fused pixel-level representation, leading to larger errors. A shallow adapter improves the results but remains weaker than the full setting. The complete 3-block residual adapter achieves the best performance, showing that feature adaptation before the PDC point classification and point regression heads improves pixel-level dense counting.

\begin{figure*}[tp]
\centering
\includegraphics[width=0.65\textwidth,keepaspectratio]{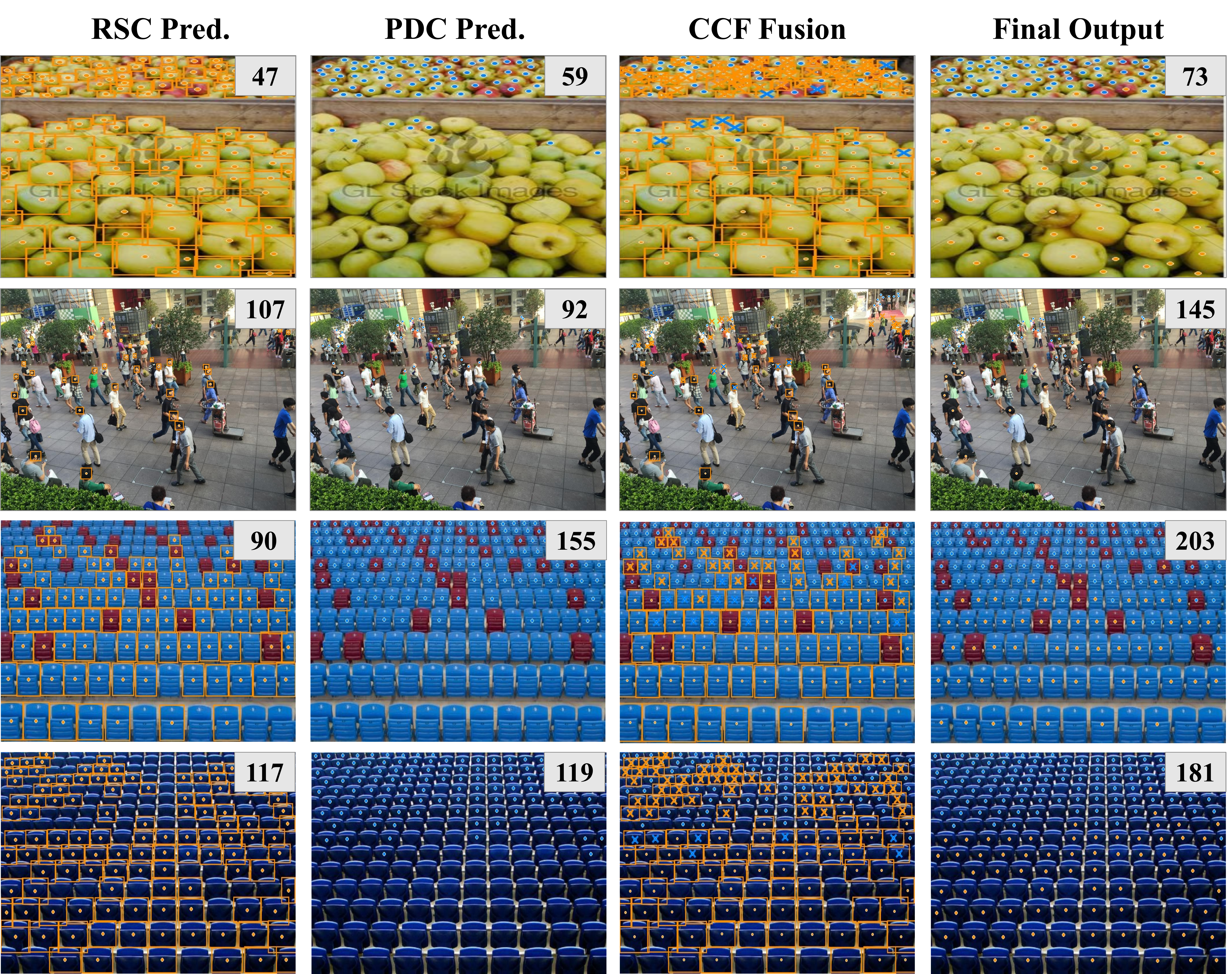}

\caption{
Qualitative visualization of Complementary Count Fusion. Orange boxes and points denote RSC predictions, while blue points denote PDC predictions. In the CCF panel, crossed markers indicate suppressed duplicate predictions between the two branches. The final output preserves complementary predictions from both branches, and the number in the top-right corner denotes the predicted count.
}
\label{fig:app_ccf_visualization}
\vspace{-10pt}
\end{figure*}

\begin{table}[!t]

\centering
\small
\setlength{\tabcolsep}{3pt}
\renewcommand{\arraystretch}{1.15}

\caption{Sensitivity analysis of the local duplicate-group size in the CCF.}
\label{tab:ablation_ccf_budget}

\begin{tabularx}{\columnwidth}{@{}cYrr@{}}
\toprule
\multirow{2}{*}{\textbf{$K_s$}} &
\multirow{2}{=}{\centering\textbf{Local duplicate group}} &
\multicolumn{2}{c}{\textbf{Metric}} \\
\cmidrule(lr){3-4}
& & \textbf{MAE} & \textbf{RMSE} \\
\midrule
0 & direct union & 22.85 & 50.05 \\
\textbf{1} & \textbf{RSC + nearest PDC candidate} & \textbf{9.34} & \textbf{33.34} \\
2 & RSC + 2 nearest PDC candidates & 9.50 & 34.26 \\
4 & RSC + 4 nearest PDC candidates & 9.75 & 36.93 \\
all & RSC + all PDC candidates inside the region & 9.85 & 37.93 \\
\bottomrule
\end{tabularx}
\end{table}
\paragraph{CCF local duplicate-group size.}
Table~\ref{tab:ablation_ccf_budget} analyzes the effect of the local duplicate-group size in CCF. This ablation generalizes the default nearest-PDC comparison used in CCF. We denote by $K_s$ the maximum number of nearest PDC candidates grouped with each retained RSC candidate inside its region; the default CCF in our method exactly corresponds to $K_s=1$. When $K_s=0$, the two-branch outputs are directly merged, keeping duplicate predictions from RSC and PDC for the same target. When $K_s>0$, each retained RSC candidate forms a local duplicate group with its $K_s$ nearest PDC candidates inside the region, and only the highest-confidence candidate in the group is kept. As $K_s$ increases beyond $1$, more nearby PDC candidates are included in the same duplicate group, which can remove useful dense-branch predictions in crowded regions and gradually increases counting errors. When $K_s=\mathrm{all}$, all PDC candidates inside each RSC region are included in the group. The setting $K_s=1$ achieves the best performance, showing that grouping only the nearest PDC candidate better balances duplicate-count suppression and dense point recall.

\begin{table}[!t]
\centering
\small
\setlength{\tabcolsep}{3pt}
\renewcommand{\arraystretch}{1.15}

\caption{Sensitivity analysis of the IoM threshold.}
\label{tab:ablation_iom_threshold}

\begin{tabular*}{\columnwidth}{@{\extracolsep{\fill}}crr@{}}
\toprule
\multirow{2}{*}{\textbf{IoM threshold}}
& \multicolumn{2}{c}{\textbf{Metric}} \\
\cmidrule(lr){2-3}
& \textbf{MAE} & \textbf{RMSE} \\
\midrule
0.3 & 9.43 & 33.41 \\
\textbf{0.5} & \textbf{9.34} & \textbf{33.34} \\
0.7 & 9.55 & 33.50 \\
0.9 & 10.01 & 33.82 \\
\bottomrule
\end{tabular*}
\end{table}

\paragraph{IoM threshold.}
Table~\ref{tab:ablation_iom_threshold} studies the effect of the IoM threshold used in RSC duplicate removal. As the IoM threshold increases, two regions need a higher overlap ratio to be regarded as duplicates, weakening the removal of duplicate RSC predictions. When the IoM threshold is too low, the removal rule becomes overly permissive and may suppress nearby non-duplicate regions. The setting of $0.5$ achieves the best performance in this group of experiments and is consistent with the IoM-NMS threshold used in SAM3~\cite{sam3} object counting evaluation. Therefore, we use IoM threshold $0.5$ as the default setting in our main experiments.

\begin{table}[!t]
\centering
\small
\setlength{\tabcolsep}{3pt}
\renewcommand{\arraystretch}{1}

\caption{Sensitivity analysis of PDC candidate density.}
\label{tab:ablation_pdc_density}

\begin{tabular*}{\columnwidth}{@{\extracolsep{\fill}}crr@{}}
\toprule
\multirow{2}{*}{\textbf{Candidates per cell}}
& \multicolumn{2}{c}{\textbf{Metric}} \\
\cmidrule(lr){2-3}
& \textbf{MAE} & \textbf{RMSE} \\
\midrule
\textbf{$1\times1$} & \textbf{9.34} & \textbf{33.34} \\
$2\times2$ & 20.38 & 122.88 \\
$3\times3$ & 21.47 & 128.80 \\
\bottomrule
\end{tabular*}
\end{table}

\paragraph{PDC candidate density.}
Table~\ref{tab:ablation_pdc_density} compares different numbers of PDC candidates generated at each spatial location. This ablation corresponds to the candidate generation process on the dense spatial grid in PDC. With the input resolution of $1008\times1008$, the default $1\times1$ setting already generates PDC candidates at approximately $7\times7$ pixel intervals, providing sufficiently dense point coverage. Simply increasing the number of candidates per spatial location introduces more adjacent candidates under the same sparse-matching and loss configuration. This increases foreground-background imbalance and duplicate candidate competition, making optimization less stable and leading to larger counting errors. Therefore, we use one PDC candidate per spatial location as the default setting.

\subsection{Qualitative Experiments}

\paragraph{Visualization of Complementary Count Fusion.}
To further illustrate the interaction between the two counting branches, Figure~\ref{fig:app_ccf_visualization} visualizes the RSC predictions, PDC predictions, fusion process, and final output. RSC provides region-level object anchors for clearly bounded instances, while PDC recovers dense point-level predictions; CCF suppresses duplicated predictions and retains complementary ones to form the final point set.

\begin{figure*}[p]
    \centering
    \includegraphics[angle=270,origin=c,width=0.54\textheight,keepaspectratio]{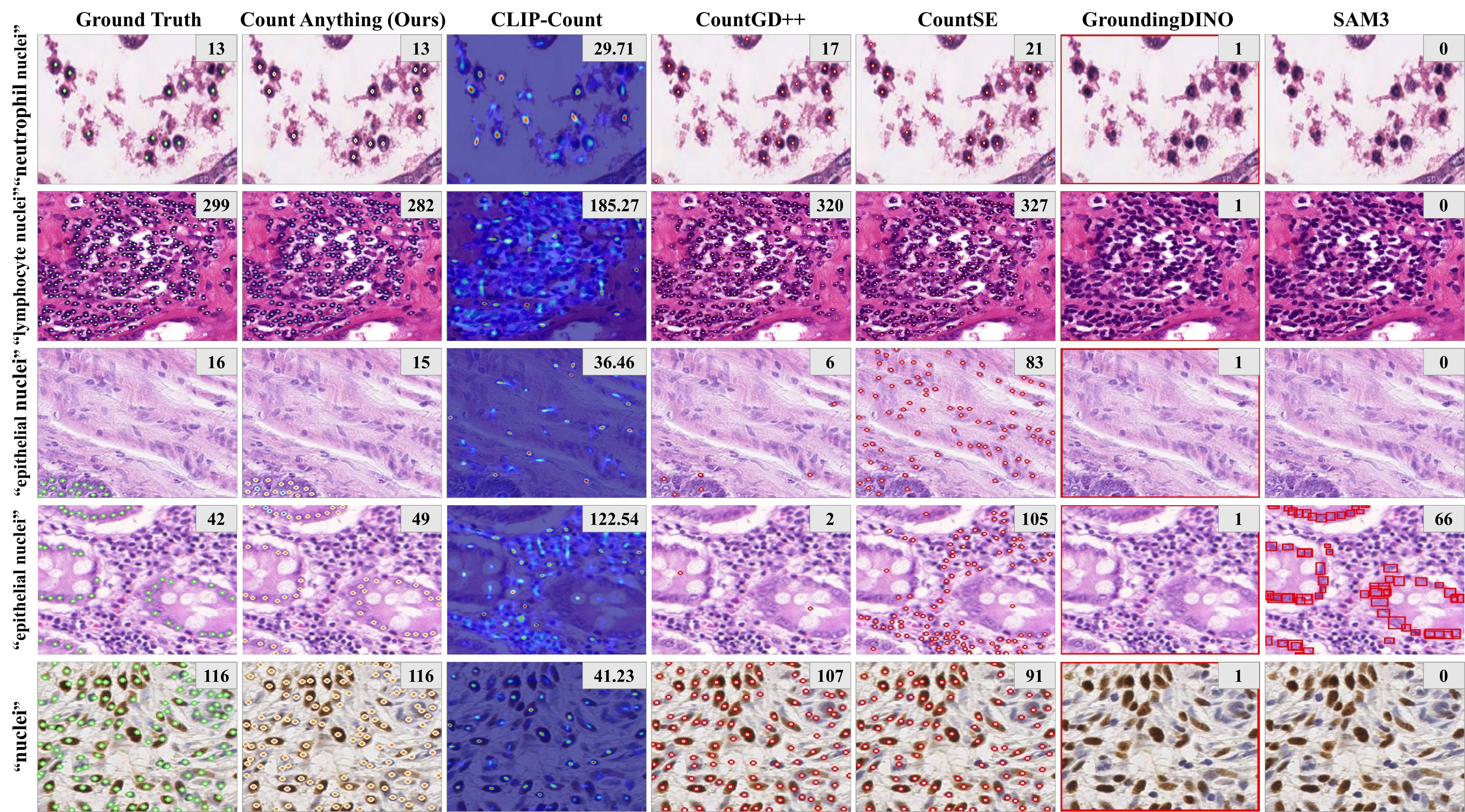}
    \caption{Additional qualitative comparisons on the Histopathology domain.}
    \label{fig:app_qual_histopathology}
\end{figure*}

\begin{figure*}[p]
    \centering
    \includegraphics[angle=270,origin=c,width=0.53\textheight,keepaspectratio]{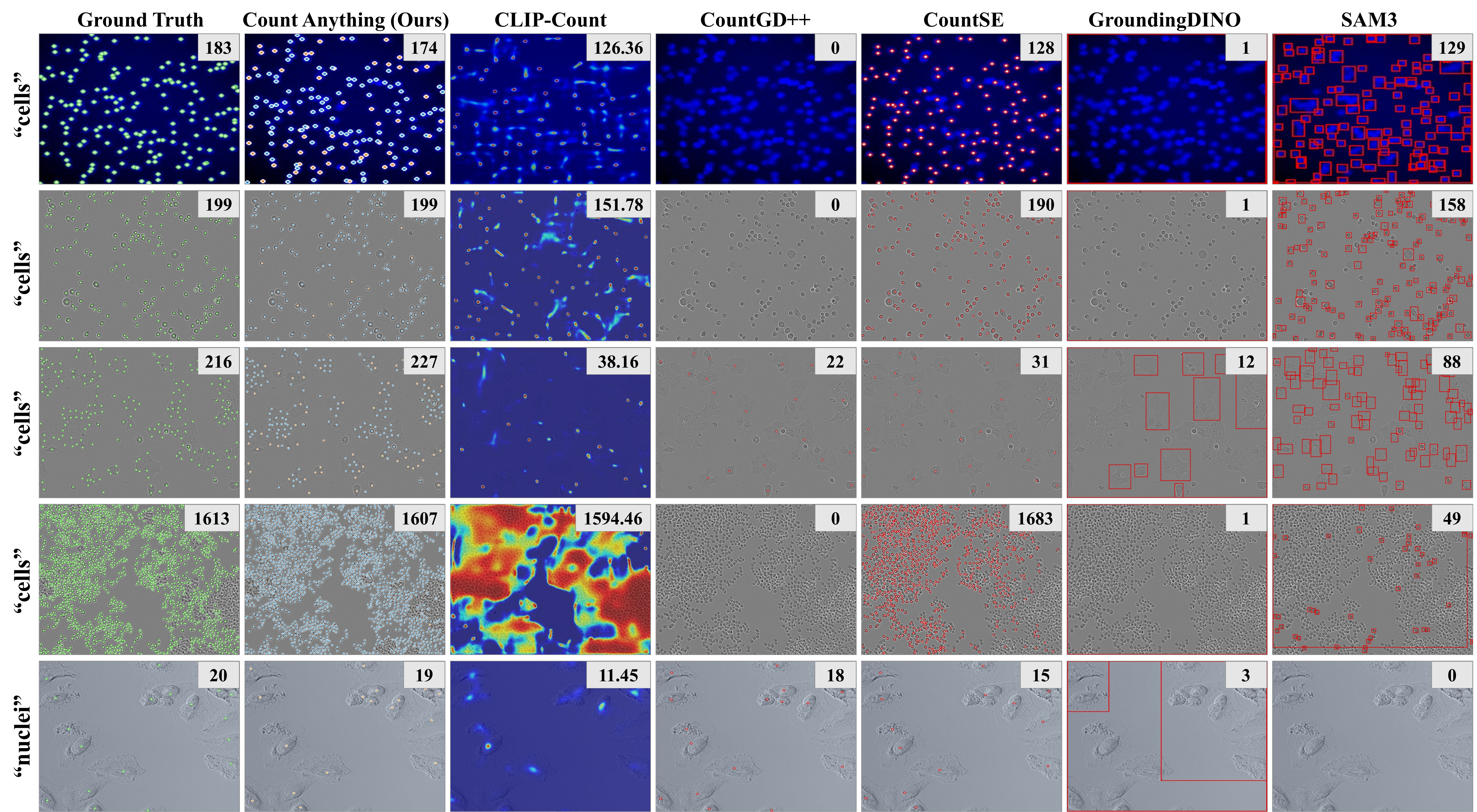}
    \caption{Additional qualitative comparisons on the Cellular Microscopy domain.}
    \label{fig:app_qual_cellular_microscopy}
\end{figure*}

\paragraph{Additional domain-wise qualitative comparisons.}
To supplement Figure~\ref{fig:quantitative_visual_comparison} in the main paper, we provide additional domain-wise qualitative comparisons on the CLOC test set in Figures~\ref{fig:app_qual_general_scene}--\ref{fig:app_qual_microbiology}, covering General Scene, Remote Sensing, Histopathology, Cellular Microscopy, Agriculture, and Microbiology. We include CountSE~\cite{countse} and GroundingDINO~\cite{grounding_dino} as compared methods.

\clearpage

\begin{figure*}[p]
    \centering
    \includegraphics[angle=270,origin=c,width=0.53\textheight,keepaspectratio]{figures/appendix/quantitative_general_scene.pdf}
    \caption{Additional qualitative comparisons on the General Scene domain.}
    \label{fig:app_qual_general_scene}
\end{figure*}

\clearpage

\begin{figure*}[p]
    \centering
    \includegraphics[angle=270,origin=c,width=0.53\textheight,keepaspectratio]{figures/appendix/quantitative_remote_sensing.pdf}
    \caption{Additional qualitative comparisons on the Remote Sensing domain.}
    \label{fig:app_qual_remote_sensing}
\end{figure*}

\clearpage

\clearpage

\clearpage

\begin{figure*}[p]
    \centering
    \includegraphics[angle=270,origin=c,width=0.54\textheight,keepaspectratio]{figures/appendix/quantitative_agriculture.pdf}
    \caption{Additional qualitative comparisons on the Agriculture domain.}
    \label{fig:app_qual_agriculture}
\end{figure*}

\clearpage

\begin{figure*}[p]
    \centering
    \includegraphics[angle=270,origin=c,width=0.53\textheight,keepaspectratio]{figures/appendix/quantitative_microbiology.pdf}
    \caption{Additional qualitative comparisons on the Microbiology domain.}
    \label{fig:app_qual_microbiology}
\end{figure*}

\clearpage

\section{Detailed Report on the CLOC Dataset}
\label{app:cloc_details}

\subsection{Introduction to the CLOC Dataset}
\label{app:cloc_motivation_overview}

\subsubsection{Motivation}

Recent progress in generalist vision models has shown that large-scale and diverse data are essential for enabling models to generalize in open-world scenarios~\cite{clip,sam}. However, in object counting, existing datasets still provide limited support for training and evaluating generalist counting models. Crowd counting datasets represented by ShanghaiTech~\cite{shanghaitech} mainly focus on counting people or heads, where the task definition is clear but the target type is limited. Category-conditioned counting datasets such as FSC-147~\cite{fsc147} expand the range of countable categories, but their image scale and visual-domain coverage remain limited. As counting tasks move toward real-world applications, models need to handle diverse visual domains such as General Scene, Remote Sensing, Histopathology, Cellular Microscopy, Agriculture, and Microbiology, as well as target-count variations from sparse objects to extremely dense scenes. Existing counting datasets remain limited in visual-domain coverage, data scale, and scene diversity, which restricts the development of generalist object counting models.

A general-purpose counting dataset for real-world scenarios should satisfy three basic requirements. First, it should cover multiple visual domains, so that models can learn counting patterns under different imaging conditions, object appearances, and scene structures, rather than adapting only to a fixed type of scenario. Second, it should have sufficient scale in terms of images, categories, and instances to support the training of generalist counting models. Finally, the dataset should establish unified annotation semantics for counting, so that each valid target can be consistently associated with one countable instance. Only with such a unified protocol can models be trained and evaluated across different categories, visual domains, and target-density ranges.

\subsubsection{Overview of CLOC}

Motivated by the above observations, we construct \textbf{CLOC}, a large-scale cross-domain object counting dataset for real-world scenarios. CLOC is built upon multiple public datasets and reorganizes their images, categories, and instance annotations around the counting task. It covers General Scene, Remote Sensing, Histopathology, Cellular Microscopy, Agriculture, and Microbiology in a unified text-guided counting setting. In this setting, category names are used as textual queries: given an image and a target query, the model is required to predict the number of instances corresponding to that query in the image. Thus, category-specified counting in CLOC serves as a concrete form of text-guided object counting.

By selecting countable instances, unifying the category space, and reorganizing counting annotations, data originally designed for detection, segmentation, crowd counting, medical image analysis, remote sensing recognition, and agricultural phenotyping are transformed into a unified data resource for training and evaluating generalist object counting models.

\begin{table}[t]
    \centering
    \caption{Statistical comparison of CLOC with representative counting and detection datasets.}
    \label{tab:dataset_comparison}

    \footnotesize
    \setlength{\tabcolsep}{1pt}
    \renewcommand{\arraystretch}{1.18}

    \begin{tabular*}{\columnwidth}{@{\extracolsep{\fill}}lrrrrrc@{}}
        \toprule
        \textbf{Dataset} &
        \textbf{Images} &
        \makecell[c]{\textbf{Cate-}\\\textbf{gories}} &
        \textbf{Instances} &
        \makecell[c]{\textbf{Instances/}\\\textbf{img}} &
        \makecell[c]{\textbf{Std.}\\\textbf{Inst./img}} &
        \makecell[c]{\textbf{Cross-}\\\textbf{domain}} \\
        \midrule
        \makecell[l]{\textbf{OmniCount-}\\\textbf{191}~\cite{omnicount191}}
        & 30,230 & 191 & 302.3K & 10.00 & 14.50 & $\times$ \\

        \textbf{FSC-147}~\cite{fsc147}
        & 6,135 & 147 & 343.8K & 56.04 & 108.35 & $\times$ \\

        \textbf{MS COCO}~\cite{mscoco}
        & 328K & 91 & 2,500K & 7.70 & 7.38 & $\times$ \\

        \textbf{VOC2007}~\cite{voc2007}
        & 9,963 & 20 & 24.6K & 2.47 & 3.09 & $\times$ \\

        \textbf{Objects365}~\cite{objects365}
        & 638K & 365 & 10,101K & 15.80 & 11.61 & $\times$ \\

        \textbf{FSCD-LVIS}~\cite{fscd_lvis}
        & 6,195 & 372 & 220.7K & 35.62 & 41.90 & $\times$ \\
        \textbf{KubriCount}~\cite{kubricount}
        & 110K & 157 & 7,320K & 66.24 & 56.66 & $\times$ \\

        \textbf{MixCount}~\cite{mixcount}
        & 58,100 & 1,522 & 3,768K & 64.85 & 34.12 & $\times$ \\

        \textbf{TPC-268}~\cite{tpc268}
        & 10,000 & 268 & 678.05K & 67.81 & 141.16 & \ding{51}(3) \\

        \midrule
        \textbf{CLOC (Ours)}
        & 220K & 619 & 15,356K & 69.74 & 187.76 & \ding{51}(6) \\
        \bottomrule
    \end{tabular*}
\end{table}

Table~\ref{tab:dataset_comparison} compares CLOC with representative counting and detection datasets. Overall, CLOC has two core characteristics: \textbf{cross-domain coverage} and \textbf{large scale}. Existing counting datasets are mostly built around a single visual source, whereas CLOC covers six visual domains and enables models to be evaluated under a unified category-specified counting setting across different visual sources. TPC-268~\cite{tpc268} also provides cross-domain coverage, but its image and instance counts are substantially lower than those of CLOC. Meanwhile, CLOC contains about $220$K images, $619$ categories, and $15.356$M instances, providing a stronger data foundation for training and evaluating generalist object counting models.

Beyond cross-domain coverage and scale, CLOC also provides a wider target-count distribution. Its higher average number of instances per image and larger standard deviation indicate that the dataset includes not only low- and medium-density counting scenes, but also high-density and extremely dense target scenes. This distribution allows models to be evaluated under different levels of counting difficulty, instead of being tested only within a fixed density range.

At the category and visual-domain levels, CLOC also preserves the complex distributions of multi-source real-world data. On the one hand, its $619$ countable categories form a large category space with a long-tailed distribution. On the other hand, its six visual domains differ substantially in image source, imaging style, and target appearance, allowing the benchmark to evaluate not only model performance in the General Scene domain, but also adaptability to Remote Sensing, Histopathology, Cellular Microscopy, Agriculture, and Microbiology. Overall, CLOC is not a specialized counting dataset for a single scenario or target type, but a large-scale cross-domain benchmark for generalist object counting.

\subsection{Dataset Statistics}
\label{app:dataset_statistics}

This section analyzes the statistical properties of CLOC from four perspectives: visual domains, categories, target counts, and image resolution. The visual-domain distribution characterizes the cross-domain coverage of CLOC and the scale differences across domains. The category distribution describes the long-tailed structure of target categories and their cross-domain composition. The target-count distribution shows the range of target counts in image-category pairs. The image-resolution distribution reflects scale differences across multi-source images. These statistics show that CLOC is diverse in visual domains, target categories, count ranges, and image scales, providing a data foundation for evaluating model generalization in complex cross-domain counting scenarios.

\subsubsection{Visual-domain Distribution}

\begin{table}[!t]
\centering
\small
\setlength{\tabcolsep}{3pt}
\renewcommand{\arraystretch}{1.15}
\caption{Image and instance distributions of CLOC across visual domains. The table reports the number of images and instances in the six visual domains of CLOC.}

\label{tab:domain_distribution}

\begin{tabular*}{\columnwidth}{@{\extracolsep{\fill}}lrr@{}}
\toprule
\textbf{Domain} & \textbf{Images} & \textbf{Instances} \\
\midrule
General Scene & 153,457 & 8,354,057 \\
Remote Sensing & 30,324 & 2,731,480 \\
Histopathology & 15,513 & 1,716,363 \\
Cellular Microscopy & 8,715 & 1,735,734 \\
Agriculture & 6,846 & 543,947 \\
Microbiology & 5,324 & 274,473 \\

\bottomrule
\end{tabular*}
\end{table}

\begin{figure*}[tp]
\centering
\includegraphics[width=0.96\textwidth,keepaspectratio]{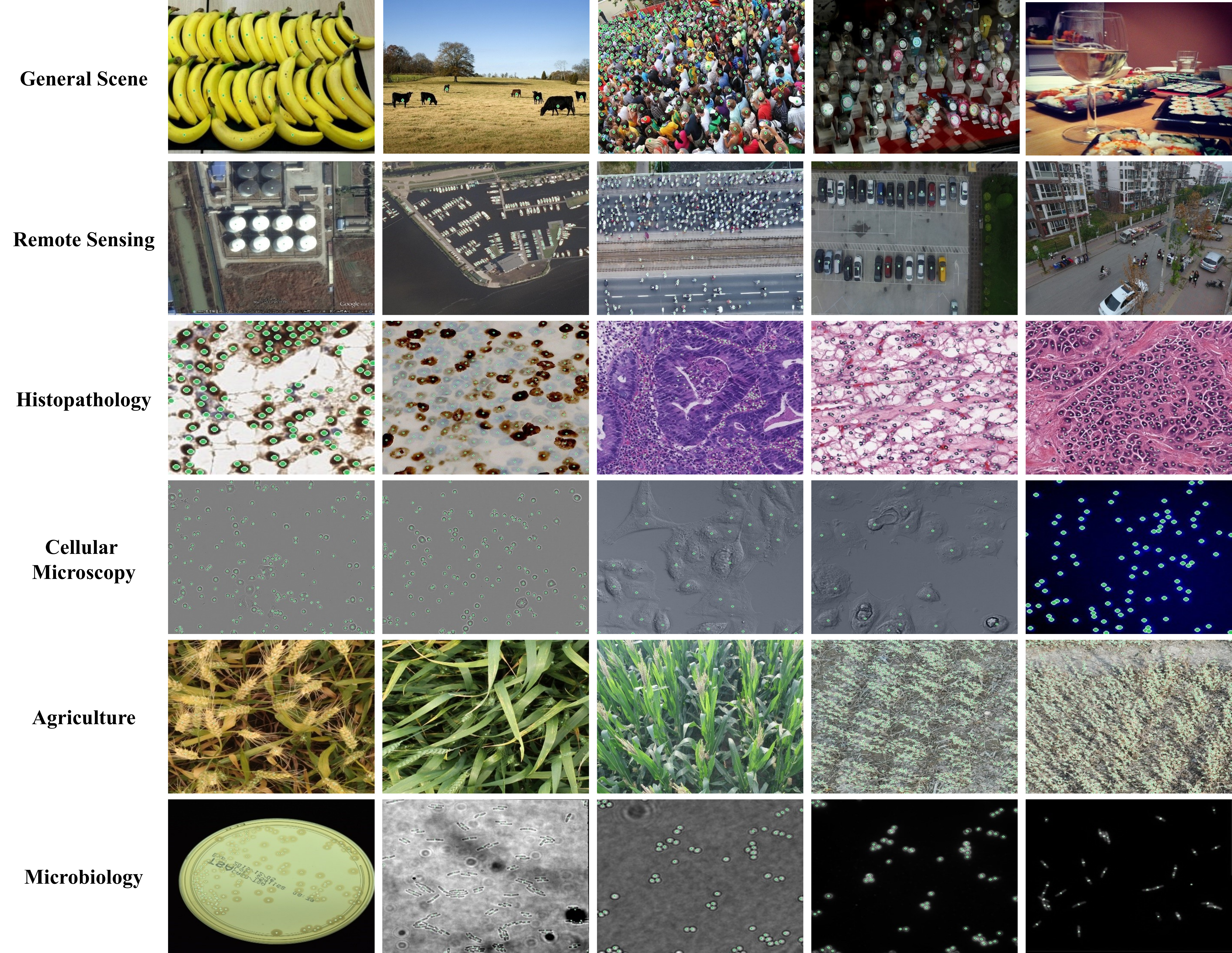}

\caption{Representative visual examples from the six visual domains of CLOC. Each row shows samples from one visual domain, including General Scene, Remote Sensing, Histopathology, Cellular Microscopy, Agriculture, and Microbiology. These examples illustrate the visual heterogeneity introduced by the multi-source construction of CLOC.}
\label{fig:domain_examples}

\end{figure*}

As shown in Figure~\ref{fig:domain_examples}, CLOC covers six visually diverse domains: General Scene, Remote Sensing, Histopathology, Cellular Microscopy, Agriculture, and Microbiology. Unlike counting datasets built from a single visual source, CLOC incorporates multiple visual domains into a unified counting benchmark, allowing models to be evaluated under the same protocol on data from different visual conditions.

In terms of sample scale, the domains are not fully balanced after being converted into the final counting framework. General Scene and Remote Sensing account for relatively larger portions of the dataset, while specialized domains such as Histopathology, Cellular Microscopy, Agriculture, and Microbiology contain fewer counting samples. This imbalance is related to the original task forms and annotation availability of public multi-source data. For data from the General Scene and Remote Sensing domains, many public datasets are released with detection, counting, or instance-level target annotations, making them easier to convert into category-specified counting supervision. For data from the Histopathology, Cellular Microscopy, Agriculture, and Microbiology domains, public datasets are also widely available, but many of them are originally designed for classification, diagnosis, segmentation, or phenotyping tasks~\cite{medmnistv2,medical_decathlon}, and do not necessarily provide complete and stable instance-level annotations that can be directly converted into counting supervision. As a result, the visual-domain scales in the final unified counting framework are not fully balanced.

It is worth emphasizing that the value of smaller visual domains does not mainly lie in their image quantity, but in the visual heterogeneity they introduce into CLOC. Remote Sensing, Histopathology, Cellular Microscopy, Agriculture, and Microbiology correspond to different imaging processes, observation scales, texture structures, and target appearances. Their counting scenarios differ substantially from conventional object counting in general-scene images. In other words, although these domains contain fewer samples, they broaden the visual coverage of the benchmark, allowing it to evaluate not only model performance in the General Scene domain, but also adaptability to Remote Sensing, Histopathology, Cellular Microscopy, Agriculture, and Microbiology. Therefore, the cross-domain property of CLOC is reflected not only in the number of covered visual domains, but also in the visual differences and counting challenges introduced by these domains.

\subsubsection{Category Distribution}

\begin{figure*}[tp]
\centering
\includegraphics[width=\textwidth,height=0.82\textheight,keepaspectratio]{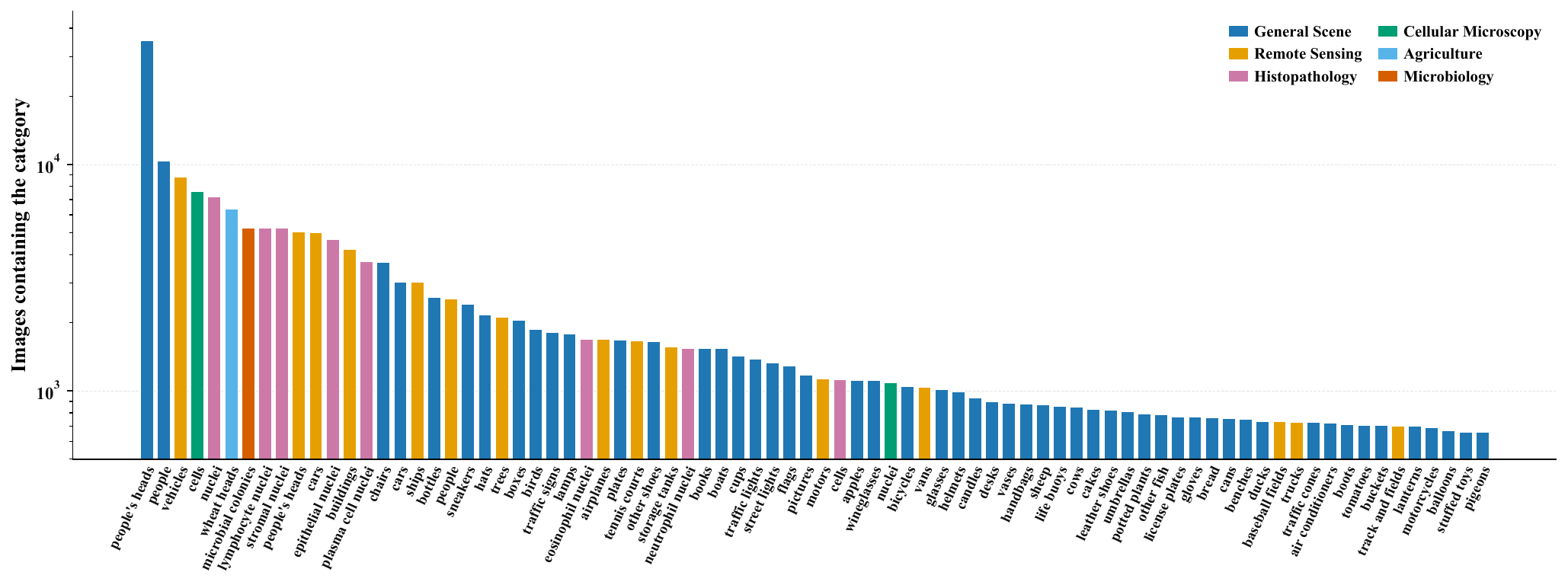}

\caption{Category-level image frequency distribution. The figure shows the top 80 category-domain pairs ranked by the number of images containing the corresponding category within each domain. The y-axis uses a logarithmic scale, and colors indicate the visual domain of each category.}
\label{fig:category_frequency}

\end{figure*}

This statistic is image-based: the value of each category-domain pair denotes how many images contain the corresponding target category within that domain. Overall, CLOC exhibits a clear long-tailed category distribution. A small number of categories have high image frequencies, while many categories gradually decrease in frequency along the ranked order. This distribution shows that CLOC is not constructed only around a few frequent targets. Instead, while covering high-frequency categories, it also includes a large number of low-frequency categories. For generalist object counting, such category coverage supports training and evaluation over a broader target-category space.

The color distribution further shows that high-frequency categories are not dominated solely by General Scene. They also include categories from the Remote Sensing, Histopathology, Cellular Microscopy, Agriculture, and Microbiology domains. In other words, the head categories of CLOC already have a cross-domain composition, rather than consisting only of common objects from the General Scene domain. This aligns with the goal of cross-domain counting: even for frequent categories, models still need to adapt to different imaging conditions, target appearances, and visual contexts.

Therefore, the category distribution supports generalist object counting evaluation in two ways. First, the long-tailed distribution expands the target-category range of the benchmark, enabling evaluation from common categories to low-frequency categories. Second, the cross-domain composition of high-frequency categories prevents models from adapting only to a single visual source, and instead requires them to learn stable counting patterns across different visual domains. This enables CLOC to provide a more comprehensive training and evaluation foundation for cross-category and cross-domain generalist counting models.

\subsubsection{Target-count Distribution}

\begin{figure*}[tp]
\centering
\includegraphics[width=\textwidth,height=0.82\textheight,keepaspectratio]{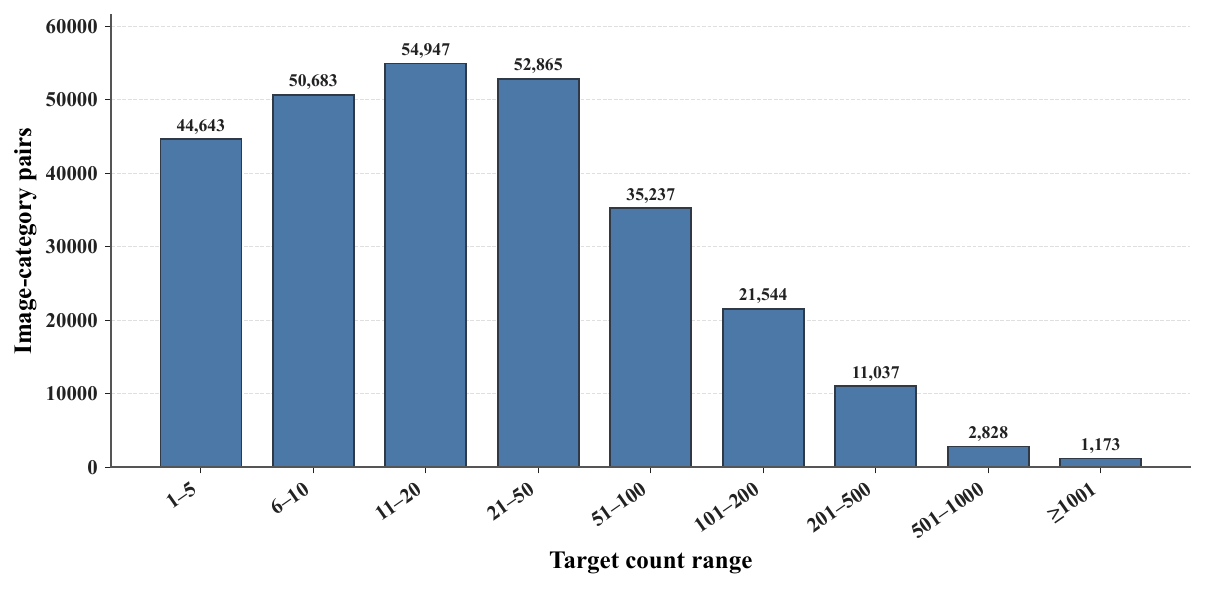}

\caption{Target-count distribution. The x-axis indicates the target-count range for each image-category pair, and the y-axis indicates the number of corresponding image-category samples.}
\label{fig:target_count_distribution}

\end{figure*}

CLOC covers a wide counting range from \texttt{1-5} to \texttt{1001+}. Low- and medium-count ranges contain a large number of samples and form the major portion of the dataset, showing that CLOC covers common counting scenarios in real-world applications. At the same time, CLOC also retains samples in higher-count ranges, enabling models to be evaluated under dense target scenes.

This distribution is consistent with the target-count structure in real-world scenarios: most image-category pairs fall into low- to medium-count ranges, while high-density and extremely dense scenes are less frequent but still present. Therefore, CLOC can evaluate not only the basic counting ability of models under common target-count ranges, but also their stability in high-count and dense target scenarios.

\subsubsection{Image-resolution Distribution}

\begin{table}[!t]
\centering
\small
\setlength{\tabcolsep}{3pt}
\renewcommand{\arraystretch}{1.15}
\caption{Mean image resolution and megapixels of CLOC across visual domains. The table reports image-scale statistics across different visual domains and shows the resolution distribution of multi-source images.}

\label{tab:domain_resolution}

\begin{tabularx}{\columnwidth}{@{}Xcr@{}}
\toprule
\textbf{Domain} & \makecell[c]{\textbf{Mean}\\\textbf{Resolution}} & \makecell[r]{\textbf{Mean}\\\textbf{Megapixels}} \\
\midrule
General Scene & $887.52 \times 705.75$ & 0.864 MP \\
Remote Sensing & $1246.06 \times 1013.23$ & 1.450 MP \\
Histopathology & $360.26 \times 357.57$ & 0.161 MP \\
Cellular Microscopy & $589.37 \times 478.80$ & 0.291 MP \\
Agriculture & $1166.15 \times 1117.30$ & 1.522 MP \\
Microbiology & $3023.68 \times 3383.06$ & 11.031 MP \\
\midrule
Overall & $941.31 \times 786.71$ & 1.125 MP \\
\bottomrule
\end{tabularx}
\end{table}

Since CLOC covers multiple visual domains, image scales vary substantially across domains. Overall, Microbiology has the highest average number of megapixels, while Agriculture and Remote Sensing also have relatively large image scales. In contrast, Histopathology and Cellular Microscopy have lower average resolutions.

These resolution differences are related to the imaging processes of different visual domains. General-scene images are usually collected by conventional cameras or public vision datasets and have diverse image sizes. Images from the Remote Sensing and Agriculture domains often cover larger spatial regions~\cite{dota_cvpr,xview,gwhd}. Images from the Microbiology domain may have higher original resolutions. To ensure that images from different sources and resolutions can be used stably within a unified training and evaluation framework, we normalize extremely high-resolution images during dataset construction, reducing the instability caused by very large images in data loading, memory consumption, and statistical distribution.

\subsection{Dataset Construction Pipeline}
\label{app:dataset_construction_pipeline}

\begin{figure}[t]
    \centering
    \includegraphics[width=\columnwidth]{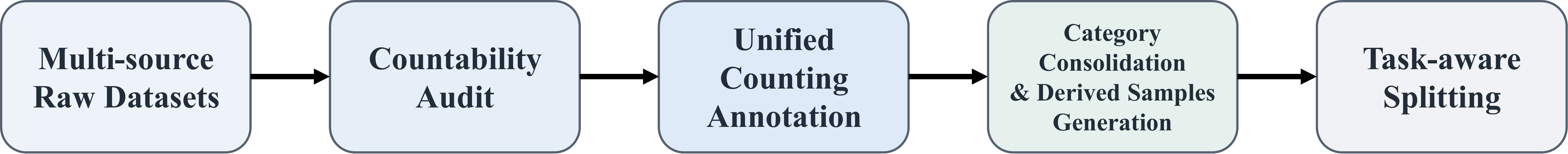}
    \caption{Overview of the CLOC dataset construction pipeline.}
    \label{fig:cloc_pipeline}
\end{figure}

Figure~\ref{fig:cloc_pipeline} illustrates the overall construction pipeline of CLOC. We first collect raw images and annotations from multiple public datasets to form a multi-source raw data pool covering six visual domains. Since these datasets were originally designed for different tasks, such as detection, segmentation, crowd counting, medical image analysis, remote sensing recognition, and agricultural phenotyping, directly merging them would introduce inconsistencies in annotation semantics, category granularity, and data splitting, as also observed in multi-dataset segmentation and detection benchmarks~\cite{mseg,bigdetection}. Therefore, the construction process is not a simple concatenation of multiple datasets. Instead, it is organized around the category-specified counting task and proceeds through countability auditing, unified counting annotation construction, category consolidation and derived-sample generation, and task-aware splitting.

Specifically, countability auditing determines whether the original annotations satisfy the one-instance-one-count requirement, and filters or skips annotation items, categories, or images that are not suitable as counting supervision. Unified counting annotation construction addresses differences in both original annotation protocols and instance representations. It converts heterogeneous annotation protocols, including COCO-style JSON, custom JSON, XML, CSV, TXT, MAT/NumPy/HDF5, NIfTI, and mask images, into a unified counting annotation structure organized by image and indexed by category. Meanwhile, different instance representations, such as bbox, point, polygon, mask, rotated box, and label map, are converted into instance-level representations suitable for counting.

Based on this unified structure, we further organize the global category space by normalizing category names, removing categories unsuitable for counting, and constructing category groups for semantically related classes. These category groups support a stricter seen / unseen category split: seen categories refer to target categories observed during training, while unseen categories are used only in validation or testing to evaluate category generalization. The category-group constraint further prevents semantically similar categories from being split across the seen and unseen sides. We then generate training-stage derived samples using two data augmentation strategies, cropping and stitching, to supplement medium- and high-count ranges and increase spatial-layout diversity. Finally, task-aware splitting constructs the training, validation, and test sets under constraints on sample source, visual-domain proportion, and category groups, forming CLOC as a multi-domain category-specified counting benchmark for training and evaluating generalist object counting models.

\subsubsection{Multi-source Raw Datasets}

\begin{table*}[!t]
\centering
\small
\setlength{\tabcolsep}{3pt}
\renewcommand{\arraystretch}{1.15}

\caption{Source datasets grouped by visual domain. The table lists the public source datasets used to construct CLOC and their corresponding visual domains.}
\label{tab:source_datasets}

\begin{tabularx}{\textwidth}{@{}>{\raggedright\arraybackslash}p{0.20\textwidth}X@{}}
\toprule
\textbf{Visual Domain} & \textbf{Source Datasets} \\
\midrule
General Scene &
Cityscapes~\cite{cityscapes}, FSC-147~\cite{fsc147}, FSCD-LVIS~\cite{fscd_lvis}, JHU-CROWD++~\cite{jhu_crowd}, NWPU-CROWD~\cite{nwpu_crowd}, Objects365~\cite{objects365}, ShanghaiTech~\cite{shanghaitech}, VOC2007~\cite{voc2007}, Rebar Counting~\cite{rebar_counting_roboflow} \\
\addlinespace[2pt]
Remote Sensing &
CARPK~\cite{carpk}, PUCPR+~\cite{carpk,pklot}, DIOR~\cite{dior}, DOTA~\cite{dota_cvpr,dota_tpami,dota_roi_transformer}, DroneCrowd~\cite{dronecrowd}, NWPU-MOC~\cite{nwpu_moc}, NWPU-VHR-10~\cite{nwpu_vhr10_part_detectors,nwpu_vhr10_survey,nwpu_vhr10_rotation}, RSOD~\cite{rsod_cnn,rsod_efhog}, VisDrone~\cite{visdrone}, UpCount~\cite{upcount}, xView~\cite{xview} \\
\addlinespace[2pt]
Histopathology &
BCData~\cite{bcdata}, CellBinDB~\cite{cellbindb}, CoNIC~\cite{conic}, EndoNuke~\cite{endonuke}, Lizard~\cite{lizard}, MoNuSAC~\cite{monusac}, MoNuSeg~\cite{monuseg_dataset,monuseg_challenge}, NuCLS~\cite{nucls}, NuInsSeg~\cite{nuinsseg} \\
\addlinespace[2pt]
Cellular Microscopy &
BriFiSeg~\cite{brifiseg}, LIVECell~\cite{livecell}, VGG~\cite{vgg_cell_counting} \\
\addlinespace[2pt]
Microbiology &
AGAR~\cite{agar}, DeepBacs~\cite{deepbacs} \\
\addlinespace[2pt]
Agriculture &
GWHD~\cite{gwhd,gwhd_2021}, MTC~\cite{mtc}, Soybean Pod Images from UAVs~\cite{soybean_pod_uav} \\
\bottomrule
\end{tabularx}

\end{table*}

As the starting point of the CLOC construction pipeline, we first collect public datasets from multiple visual domains to form a multi-source raw data pool. Here, ``raw datasets'' refer to source data before countability auditing, unified annotation conversion, category consolidation, and derived-sample generation. To enable CLOC to cover diverse real-world scenarios, we broadly collect data from different visual sources, covering the General Scene, Remote Sensing, Histopathology, Cellular Microscopy, Agriculture, and Microbiology domains, so that the subsequent construction process is built upon a richer set of visual domains.

Table~\ref{tab:source_datasets} lists the source datasets used in the current version and their corresponding visual domains. All source datasets used to construct CLOC are public datasets, and we use them under their original licenses and terms of use. We retain the original ownership, citation, license, and usage information for each source dataset in the released CLOC documentation; the derived CLOC benchmark does not override the restrictions imposed by the original assets. The raw data pool covers the six visual domains described above. Datasets from different visual domains were originally designed for different tasks, including object detection, crowd counting, instance segmentation, cell nucleus analysis, remote sensing object recognition, microorganism analysis, and agricultural phenotyping. Therefore, although these datasets contain certain forms of target annotations or instance information, their original annotation formats, category definitions, and instance granularities are not consistent, and they cannot be directly merged into a unified counting dataset.

\subsubsection{Countability Audit and Annotation Cleaning}

After collecting the multi-source raw datasets, we first determine which annotations can serve as reliable counting supervision. Different source datasets may contain special labels or regions such as \texttt{crowd}, \texttt{group}, \texttt{ignore}, \texttt{difficult}, and \texttt{NegativeROA} in their original tasks, or may contain invalid labels outside the official category space. These labels have dataset-specific meanings and cannot be simply treated as either valid or invalid counting samples. Therefore, before unified annotation conversion, we perform a countability audit according to the annotation protocol of each source dataset, and decide whether to retain, filter, mask, or skip different categories, regions, or images.

\begin{figure*}[tp]
\centering
\includegraphics[width=\textwidth,height=0.82\textheight,keepaspectratio]{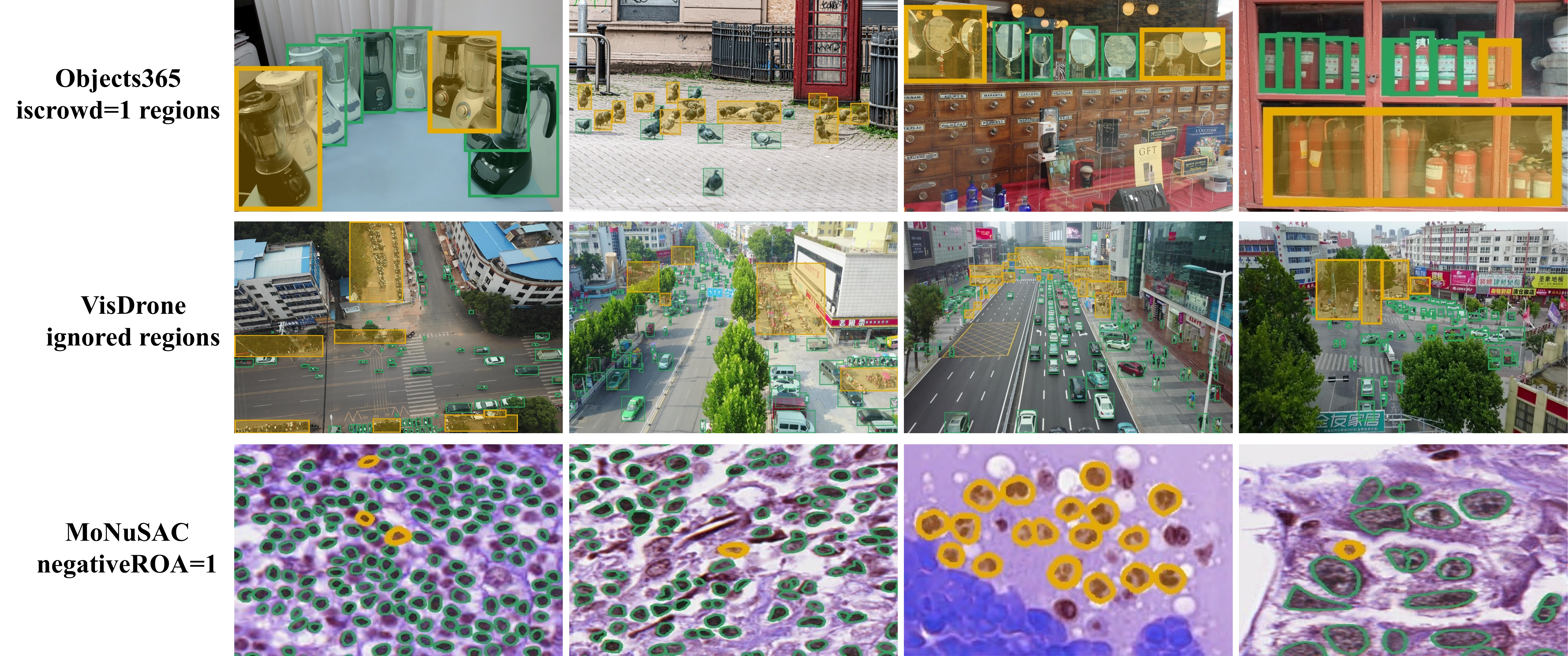}

\caption{Examples of source-specific special annotations requiring countability audit before unified conversion. Rows show \texttt{iscrowd=1} regions in Objects365, ignored regions in VisDrone, and \texttt{NegativeROA=1} regions in MoNuSAC. Green overlays denote regular instance annotations in the original datasets, while yellow overlays highlight source-specific special regions. These annotations have dataset-specific semantics and cannot be directly converted into one-instance-one-count supervision without source-specific handling.}
\label{fig:countability_audit_examples}

\end{figure*}

Figure~\ref{fig:countability_audit_examples} provides representative examples of such source-specific annotations. These cases motivate the source-aware auditing rules described below, where different special labels are handled according to their original dataset protocols rather than being uniformly retained or discarded.

\paragraph{For general-scene data} Objects365~\cite{objects365} adopts COCO-style annotations and contains \texttt{iscrowd=1} annotation items. These annotations usually represent crowded regions or target groups that cannot be separated instance by instance, where a single \texttt{crowd} annotation may correspond to multiple real instances. If such an annotation is directly counted as one instance, the target count would be underestimated; if only other regular instances in the same image are retained, targets inside the \texttt{crowd} region may still be missing. Therefore, for images containing \texttt{crowd} annotations, we do not use the corresponding category in that image as valid counting supervision. Cityscapes~\cite{cityscapes} contains \texttt{persongroup}, \texttt{cargroup}, and \texttt{bicyclegroup} labels, which similarly denote a group of objects rather than clear instance-level annotations. Thus, when a regular instance category and its corresponding \texttt{group} label coexist in the same image, we ignore the annotations of that category in the image to avoid incomplete counting ground-truth.

The \texttt{distractor} label in JHU-CROWD++~\cite{jhu_crowd} represents a different case. For images with \texttt{distractor = 1}, the label indicates complex scenes with no obvious crowd or only very few people, rather than instance-inseparable or incomplete annotations like \texttt{group} or \texttt{iscrowd}. Therefore, we retain JHU-CROWD++ images with \texttt{distractor = 1}, so that CLOC includes such low-count or near-negative crowd scenes.

\paragraph{For remote-sensing data} VisDrone~\cite{visdrone} contains \texttt{ignored region} annotations. These regions indicate image areas ignored during evaluation, where targets are not annotated, and the specific categories inside the regions cannot be reliably determined. If other targets in the same image were directly retained, unannotated instances inside the \texttt{ignored region} would make the counting supervision uncertain. Therefore, we skip images containing \texttt{ignored region}. xView~\cite{xview} has a different issue: although its official category list contains 60 valid target categories, its raw ground-truth annotations may include additional \texttt{type\_id}s that are not included in the official category list. These annotations do not belong to the official valid target categories and are excluded from scoring in the official evaluation. Therefore, we treat them as invalid annotations and filter them during data conversion, so that they do not enter subsequent training, evaluation, or point / bbox statistics.

For sequence datasets with strong temporal continuity, such as DroneCrowd~\cite{dronecrowd} and UpCount~\cite{upcount}, adjacent frames are highly similar. Keeping all frames would introduce redundant near-duplicate samples and may cause potential leakage in later data splits. Therefore, we apply fixed-interval frame sampling and retain one frame every 10 frames.

\paragraph{For medical and cellular-microscopy data} MoNuSAC~\cite{monusac} annotations are stored in Aperio ImageScope XML format and may contain regions with \texttt{NegativeROA=1}. This flag corresponds to regions drawn with the Negative Pen in ImageScope, which are intended to exclude areas from analysis rather than represent nucleus instances. Therefore, regions with \texttt{NegativeROA=1} are not converted into valid counting instances; they are treated as excluded regions during annotation conversion. In NuCLS~\cite{nucls}, the \texttt{AMBIGUOUS} class is also not treated as a stable semantic category. It usually corresponds to nucleus-like objects whose category cannot be reliably determined, whose annotation agreement is low, or for which reliable classification labels should not be provided. It is therefore different from well-defined nucleus categories such as \texttt{tumor} or \texttt{lymphocyte}. Accordingly, we do not include \texttt{AMBIGUOUS} as a valid counting category in the unified category space.

In summary, before unified annotation conversion, we first audit and clean the original annotation semantics of each data source instead of directly converting all annotations into counting supervision. Through this process, only annotations that can be stably associated with independent instances enter the subsequent data pool. Special labels or regions such as \texttt{crowd}, \texttt{group}, \texttt{ignore}, and \texttt{NegativeROA}, invalid categories, excluded regions, near-duplicate sequence frames, and categories or regions unsuitable for counting are retained, filtered, masked, or skipped according to the protocol of each dataset. This process ensures that data from different sources have as consistent a one-instance-one-count semantics as possible after entering the unified format, and reduces the impact of original annotation differences on training and evaluation.

\subsubsection{Unified Counting Annotation Construction}

After countability auditing and annotation cleaning, we further convert the original annotations from different sources and protocols into a unified counting annotation structure. Since these datasets were originally designed for different tasks, such as detection, segmentation, crowd counting, medical image analysis, remote sensing object recognition, and agricultural phenotyping, their original annotations differ in file format, field semantics, indexing scheme, and instance organization. To make these data trainable, evaluable, and reusable as a unified counting benchmark in CLOC, we convert all data sources into an image-organized and category-indexed counting annotation structure. In this structure, the target categories and instance annotations in each image can be read and used in a consistent manner. This unified structure also provides the basis for subsequent category consolidation, statistical analysis, derived-sample generation, and train/validation/test splitting.

\paragraph{1) Parsing JSON-based annotation protocols} The original datasets vary substantially in their annotation protocols. Some source datasets adopt standard COCO-style JSON annotations, such as Objects365-2020, FSCD-LVIS, and LIVECell. These annotations are typically organized around images, instance annotations, and category tables, where each instance is described by an image ID, category ID, bounding box, and related fields. Other datasets also use JSON files but do not follow the COCO schema. For example, FSC-147~\cite{fsc147} stores point annotations, data splits, and category information in separate files; LabelMe-style datasets such as Soybean Pod Images from UAVs~\cite{soybean_pod_uav} describe targets through \texttt{shapes}, \texttt{shape\_type}, and \texttt{points}; Cityscapes~\cite{cityscapes} uses polygon JSON files to record object contours and category labels; and AGAR~\cite{agar} adopts a customized sample-description format. Although these datasets all store annotations in JSON format, their field meanings, indexing schemes, and instance organizations are different, and therefore need to be parsed separately before being written into the unified structure.

\paragraph{2) Parsing non-JSON annotation protocols} Beyond JSON annotations, we also process a variety of non-JSON annotation protocols. VOC-style datasets use XML files to record the target category, bounding box, \texttt{difficult} flag, and related fields for each image. Medical datasets such as MoNuSeg and MoNuSAC also use XML files, but their annotations record polygon vertices rather than standard detection boxes. GWHD, NuCLS, and related datasets store annotations in CSV files or tabular formats, where target boxes, categories, or patch metadata need to be parsed from table fields. VisDrone, JHU-CROWD++, UpCount, CARPK, and PUCPR+ use plain text files to record target boxes, point annotations, or image-level labels. Some datasets come from scientific computing or medical imaging formats, such as \texttt{.mat} files in ShanghaiTech, DroneCrowd, NWPU-CROWD, and MTC; \texttt{.npy} arrays in CoNIC; HDF5 files in BCData; \texttt{.nii.gz} label maps in BriFiSeg; and TIF or mask-image annotations in CellBinDB, NuInsSeg, and DeepBacs. For datasets such as VGG, annotations may even exist as point maps, where counting points need to be extracted from image pixels.

\paragraph{3) Unified instance representation} During conversion, different types of instance annotations are unified into two basic representations: \texttt{point} and \texttt{bbox}. For instances that already provide bounding boxes, we retain their axis-aligned bounding boxes and generate corresponding counting points from the box centers. For rotated-box annotations, we first convert each rotated box into its horizontal enclosing box, and then generate the center point from that box. For region-level annotations such as polygons, masks, or label maps, we extract the enclosing bounding box from each instance region and further generate the corresponding center point. For datasets that only provide point annotations, we retain the original points as counting supervision and leave the bounding box field empty. Through this conversion, bbox, rotated box, polygon, mask, label map, and point annotations from different sources are unified into a common instance representation for counting.

In summary, the core of unified counting annotation construction is to convert heterogeneous protocols, including COCO JSON, custom JSON, XML, CSV, TXT, MAT/NumPy/HDF5, NIfTI, and mask images, into an image-organized and category-indexed counting data structure. This structure preserves valid instance information from different sources and enables CLOC to be released and used in a unified manner. For users, it removes the need to adapt to the original annotation protocol of each source dataset separately, allowing direct training, evaluation, and statistical analysis on the unified structure.

\begin{lstlisting}[
  style=jsonbox,
  float=t,
  caption={Example of the unified counting annotation structure. This example shows how fields such as image path, source dataset, category set, instance point annotations, bounding box annotations, sample status, and visual domain are organized in the unified annotation format.},
  label={lst:annotation_example}
]
{
  "35439": {
    "idx": 35439,
    "image_path": ".../objects365_v1_00000000.jpg",
    "image_from": "objects365",
    "classes": ["people", "projectors"],
    "annotation": {
      "people": {
        "point": [[61.6, 283.0], [121.3, 252.9]],
        "bbox": [[20.3, 260.2, 102.9, 305.7],
                 [102.9, 235.1, 139.7, 270.6]]
      },
      "projectors": {"point": [...], "bbox": [...]}
    },
    "status": "original",
    "modality": "General Scene"
  }
}
\end{lstlisting}

\subsubsection{Category Consolidation and Derived Sample Generation}

After countability auditing, annotation cleaning, and unified counting annotation construction, we further organize the category space in the global data pool and generate derived samples for training. Although different data sources have already been converted into a unified counting annotation format, they were originally designed for different tasks and annotation systems. As a result, category names, category granularity, and category hierarchies may still differ. If merged directly, the same semantic target could be split into multiple categories due to differences in capitalization, singular/plural forms, or naming conventions. The data pool may also contain categories that are not suitable as instance-level counting targets. In addition, the relationship between fine-grained subclasses and coarse-grained parent classes can affect the seen / unseen category split. Therefore, after unified annotation conversion, we further perform global category consolidation. Based on this consolidated data pool, we then generate training-stage derived samples through cropping and stitching, enriching medium- and high-count training samples and increasing the diversity of target spatial layouts.

\paragraph{1) Category name normalization} We normalize category names so that categories from different data sources use consistent written forms. Public datasets often adopt different naming conventions, such as differences in capitalization, singular/plural forms, spaces, underscores, and special characters. For such cases, we convert category names into normalized forms and merge clearly synonymous or near-synonymous expressions when necessary. For example, \texttt{Bottle} in Objects365, \texttt{bottle} in FSCD-LVIS, and \texttt{bottles} in FSC-147 are unified as \texttt{bottles}; similarly, \texttt{Chair} in Objects365, \texttt{chair} in FSCD-LVIS, and \texttt{chairs} in FSC-147 are unified as \texttt{chairs}. This process reduces category fragmentation caused by naming differences and ensures that the same target concept has a consistent category name in the global data pool.

\paragraph{2) Removing categories unsuitable for counting} We further remove categories that are not suitable as instance-level counting targets. Not all original categories in multi-source datasets can be stably associated with the ``one instance, one count'' assumption. Some categories have ambiguous boundaries, some are closer to regions or scene structures, and some may be valid for detection or recognition in their original tasks but are not suitable as category-specified counting targets. For example, in remote sensing data, categories such as \texttt{harbor}, \texttt{Expressway-Service-area}, and \texttt{overpass} usually refer to large scene structures or region-like targets, whose instance boundaries and counting units are less clear than objects such as vehicles, airplanes, or ships. These categories are therefore removed from the valid counting category space, ensuring that the retained categories can more stably support instance-level counting.

\paragraph{3) Semantic category group construction} We construct semantic category groups for some fine-grained categories to support the later seen / unseen split. For category consolidation in a multi-source dataset, it is not appropriate to simply promote all subclasses to their parent class. For example, Objects365 may contain fine-grained vehicle categories such as \texttt{Formula 1 cars}, \texttt{sports cars}, \texttt{suvs}, and \texttt{vans}. If all these categories were directly merged into \texttt{cars}, we would have to assume that all \texttt{cars} in the corresponding images are completely annotated. However, in the original detection datasets, an image may still contain other vehicle instances that are not annotated as any of these fine-grained categories. Directly merging these fine-grained labels into \texttt{cars} could therefore turn some unannotated car instances into missing labels, introducing false-negative noise.

To avoid this issue, we do not simply promote subclasses to parent classes. Instead, we construct category groups to capture semantic relationships among related categories. For example, \texttt{cars}, \texttt{sports cars}, \texttt{Formula 1 cars}, \texttt{suvs}, and \texttt{vans} are assigned to the same vehicle-related category group. Similarly, fine-grained categories related to bottles, chairs, birds, and boats are also grouped accordingly. These category groups are mainly used in later data splitting, ensuring that seen / unseen splits do not separate highly related categories across the two sides. This leads to a stricter evaluation of the model's ability to generalize to genuinely unseen categories.

\paragraph{4) High-resolution image normalization} After category consolidation, some extremely high-resolution images need to be normalized in scale. Since the dataset covers multiple visual domains, image resolutions vary substantially across data sources. For example, DOTA~\cite{dota_cvpr,dota_tpami,dota_roi_transformer}, xView~\cite{xview}, and JHU-CROWD++~\cite{jhu_crowd} contain high-resolution images. If such images were directly used in the unified training and visualization pipeline, they would introduce large memory, loading, and rendering costs, and would also create excessive scale differences across data sources. Therefore, extremely high-resolution images are cropped into image tiles whose long side does not exceed 2048, while the corresponding instance \texttt{point} and \texttt{bbox} annotations are updated accordingly. This step allows high-resolution images to enter the unified data pool at a more stable scale, while preserving their local target distributions and dense-scene information.

In the unified CLOC data pool, low- and medium-count samples are already well covered. To further enrich the medium- and high-count ranges during training, we generate two types of derived samples: cropped samples and stitched samples. Cropping mainly extracts local target regions from high-density images to produce medium- or high-count local training samples. Stitching mainly combines low- or medium-count image patches into higher-count training samples, thereby expanding the target-count range and increasing spatial-layout diversity.

\paragraph{5) Cropped sample generation} Cropped samples are generated by extracting local target regions from original images. Unlike simple random cropping, our cropping process prioritizes local windows whose target counts fall into predefined count ranges, thereby supplementing medium- and high-count ranges and local dense scenes in a targeted manner. To avoid generating a large number of highly overlapping near-duplicate crops from the same image, we impose an overlap constraint on selected windows by limiting the IoU between a new window and existing cropped regions. In the unified data pool, we use multi-scale candidate windows for local cropping, including $384\times384$, $512\times512$, $768\times768$, $384\times256$, $512\times384$, and $768\times512$. Cropped samples smaller than $384\times384$ are padded to satisfy the minimum input size.

\paragraph{6) Stitched sample generation} Stitched sample generation also follows a target-count-oriented strategy. We first select candidate image-patch combinations according to predefined target-count ranges, so that the stitched samples can supplement specified medium- and high-count intervals. Unlike direct random composition, the image patches used for stitching are required to come from the same visual domain and correspond to the same counting category. This avoids forcibly mixing samples with substantially different imaging styles and ensures consistency in both visual domain and category semantics. To reduce near-duplicate derived samples, we also limit the maximum reuse count of each patch during stitching.

\begin{figure*}[tp]
\centering
\includegraphics[width=\textwidth,height=0.82\textheight,keepaspectratio]{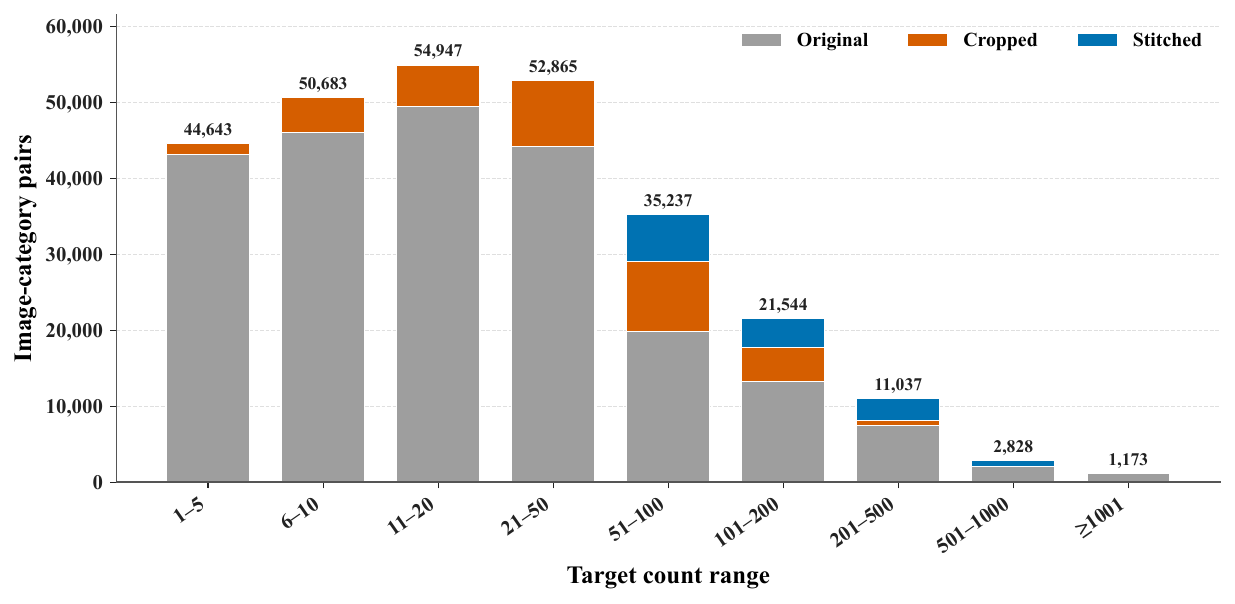}

\caption{Distribution of original and derived samples across target-count ranges. The figure shows the contribution of original samples, cropped samples, and stitched samples to each target-count range. Cropped and stitched samples are used only in the training set.}
\label{fig:derived_sample_distribution}

\end{figure*}

Figure~\ref{fig:derived_sample_distribution} shows the distribution of original samples, cropped samples, and stitched samples across different target-count ranges. The overall distribution is still mainly supported by original samples, while cropped and stitched samples provide additional coverage in medium- and high-count ranges such as \texttt{51-100}, \texttt{101-200}, and \texttt{201-500}. In other words, derived samples are used to enhance density coverage during training based on the original data, allowing the model to observe more locally dense and high-count scenes. Cropped and stitched samples are used only in the training set and are not included in the validation or test sets. The validation and test sets are composed of original samples, ensuring that CLOC evaluation reflects counting ability under the real data distribution.

\subsubsection{Task-aware Splitting}
\label{app:task_aware_splitting}

After unified annotation construction, category consolidation, and derived-sample generation, we further construct the training, validation, and test sets. Since the data pool covers multiple visual domains, contains a large number of counting categories, and includes both original and derived samples, simple random splitting cannot satisfy the evaluation requirements of a general counting benchmark. Specifically, the split needs to avoid data leakage caused by derived samples such as cropped and stitched samples, ensure that different visual domains remain evaluable in the validation and test sets, and establish a reliable seen / unseen category-generalization setting. In particular, for visual domains with a rich category space, merely ensuring that training and test category names do not overlap is not sufficient to fully characterize category generalization. If semantically related categories are split across the seen and unseen sides, the model may still benefit from strong priors learned from neighboring categories during training, thereby weakening the strictness of unseen evaluation. Therefore, we adopt a task-aware splitting strategy that constrains the final split from three aspects: derived-sample isolation, visual-domain proportions, and semantic category groups.

\paragraph{1) Derived-sample isolation} During task-aware splitting, cropped and stitched samples generated in the preceding stage to enrich the target-count distribution are retained only in the training set and excluded from the validation and test sets. All derived training samples are generated exclusively from original images assigned to the training split, ensuring that no validation or test image is used for their generation.  Meanwhile, cropped and stitched samples are used only to enhance target density, count range, and spatial-layout diversity during training. The validation and test sets are entirely based on unaugmented original samples, so that the evaluation results reflect the model's counting ability under the real image distribution.

\paragraph{2) Visual-domain proportion constraint} The split also explicitly considers the distribution of the six visual domains. To ensure stable cross-domain evaluation, the training, validation, and test sets all cover every visual domain. Under the constraints of derived-sample isolation and category splitting, we further make the image proportion of each visual domain in each split as close as possible to its proportion in the overall data pool. Compared with simple random splitting, this strategy avoids small specialized domains from being randomly weakened or omitted in the validation or test sets, allowing model performance to be jointly evaluated under different visual conditions.

\paragraph{3) Strict seen / unseen splitting based on category groups} For visual domains with rich category spaces, such as General Scene and Remote Sensing, we construct a seen / unseen evaluation setting to assess category generalization. Existing category-conditioned counting datasets usually construct unseen-category evaluation by ensuring that category names do not overlap. For example, the original FSC-147~\cite{fsc147} split ensures that the train, validation, and test sets do not share object categories, with 89 categories used for training, 29 for validation, and 29 for testing. However, later revisions of FSC-147~\cite{fsc133} point out that category-name disjointness alone can still lead to semantic split overlap. For example, ambiguous or hierarchically related categories such as \texttt{kidney beans} / \texttt{red beans} and \texttt{bread rolls} / \texttt{baguette rolls} can still create strong semantic associations between training and testing~\cite{class_agnostic_counting_survey}. This suggests that, in a large category space, seen / unseen splitting should not rely only on whether category names are identical, but should also consider semantic relationships among categories.

This issue becomes more prominent in a multi-source dataset. Different data sources often contain both coarse-grained and fine-grained categories, such as \texttt{cars}, \texttt{sports cars}, \texttt{Formula 1 cars}, \texttt{suvs}, and \texttt{vans}. If \texttt{sports cars} or \texttt{vans} appear in the training set while \texttt{cars} is placed in the test set as an unseen category, the model has in fact already observed visual concepts that are highly related to the test category during training. In this case, unseen results may reflect transfer from neighboring categories rather than true generalization to genuinely unseen semantic categories.

To mitigate this issue, we perform seen / unseen splitting based on the category groups constructed during category consolidation. Specifically, categories that are semantically close, have parent-child relationships, or can easily provide substitute supervision are assigned to the same category group, such as vehicle-related, bird-related, bottle-related, and boat-related groups. During splitting, the same category group is not divided across the seen and unseen sides. In other words, if a category group is selected as unseen, semantically related categories in that group are also excluded from the training categories or constrained accordingly; if a category group is retained as seen, its related categories will not simultaneously appear as unseen categories for generalization evaluation. In this way, the seen / unseen boundary is no longer determined only by individual category names, but is jointly constrained by semantic category groups, leading to a stricter category-generalization evaluation.

This category-group-based splitting strategy has two advantages. First, it reduces implicit leakage caused by semantically neighboring categories, making unseen evaluation closer to a genuinely unseen-category setting. Second, it preserves fine-grained categories themselves rather than simply promoting subclasses to parent classes, thereby avoiding new missing-label risks introduced by coarse-grained merging. For domains with relatively small and semantically concentrated category spaces, such as Histopathology, Cellular Microscopy, Agriculture, and Microbiology, we do not additionally construct seen / unseen category splits in the current version. These domains are mainly used to evaluate counting stability across different specialized visual domains.

The split statistics of \textbf{CLOC} are shown in Table~\ref{tab:split_statistics}. The training set contains $207.3$K images, $586$ categories, and $14.70$M instances; the validation set contains $4.7$K images, $77$ categories, and $278.4$K instances; and the test set contains $8.1$K images, $78$ categories, and $382.3$K instances. All three splits cover the six visual domains. Meanwhile, cropped and stitched samples appear only in the training set, which prevents data leakage while supporting cross-domain counting evaluation and stricter category-generalization evaluation.

\begin{table}[!t]

\centering
\small
\setlength{\tabcolsep}{3pt}
\renewcommand{\arraystretch}{1.15}
\caption{Statistics of the training, validation, and test splits. The table reports the number of images, categories, and instances in each split.}

\label{tab:split_statistics}

\begin{tabular*}{\columnwidth}{@{\extracolsep{\fill}}lccc@{}}
\toprule
 & \textbf{Train} & \textbf{Validation} & \textbf{Test} \\
\midrule
Images & 207.3K & 4.7K & 8.1K \\
Categories & 586 & 77 & 78 \\
Instances & 14.70M & 278.4K & 382.3K \\
\bottomrule

\end{tabular*}
\end{table}

\section{Extended Related Work}
\label{sec:extended_related_work}

\subsection{Object Counting Representations}

Object counting has developed several prediction representations that capture different aspects of instance distribution and spatial structure.
Density-estimation methods predict continuous response fields whose integrals provide object counts.
MCNN~\cite{shanghaitech}, CSRNet~\cite{csrnet}, and SANet~\cite{sanet} develop multi-scale processing, contextual modeling, and feature aggregation to accommodate variations in object size and crowding.
Bayesian Loss~\cite{bl} and DM-Count~\cite{dmcount} further improve how point annotations supervise density prediction.
This line of work establishes the importance of preserving local spatial information and learning effective supervision for densely distributed targets.

Detection-based counting provides explicit regions associated with individual objects.
LSC-CNN~\cite{lsc_cnn}, LPN~\cite{carpk}, and Counting-DETR~\cite{fscd_lvis} demonstrate region-based counting in crowd, vehicle, and category-agnostic settings.
Region predictions encode object extent as well as location, providing geometric structure for instance enumeration.
Point-regression methods establish a direct connection between counting and instance localization.
P2PNet~\cite{p2pnet} predicts instance points and confidence scores through one-to-one matching, and subsequent work further develops point-based localization and supervision~\cite{apgcc,caapn}.
Point predictions provide discrete counting units without requiring complete object boundaries, supporting localization in dense and weakly bounded scenes.

These developments identify object-level spatial structure and dense-instance localization as important capabilities for reliable counting.
Their relevance varies with object scale, crowding, and boundary clarity, all of which change substantially across visual domains.
However, density-only outputs lack explicit instance correspondence, region-based predictions may miss densely packed targets, and point-only predictions provide limited geometric information about object extent.
Count Anything brings region-level anchoring and pixel-level localization into a common instance-point representation.
Its point-centric supervision connects region and point predictions to the same counting instances, while complementary fusion reduces duplicate predictions and preserves distinct instances in crowded regions.

\subsection{Generalist Object Counting}

Category-agnostic and few-shot counting broaden the target space beyond categories fixed during model training.
FSC-147 and FamNet~\cite{fsc147} provide an influential benchmark and model for counting diverse object categories, while CounTR~\cite{countr} develops Transformer-based representations for generalized counting.
Reference-free methods also explore the use of repetitive visual patterns for class-agnostic counting~\cite{fsc133}.
Together, these studies establish transferable visual representations and target-conditioned matching as foundations for generalist counting.

Text-guided methods connect counting with semantic target specification.
CLIP-Count~\cite{clip_count} and VLCounter~\cite{vlcounter} exploit vision-language representations, while T2ICount~\cite{t2icount} investigates generative priors for text-guided counting.
Candidate-based methods further develop explicit instance reasoning: DAVE~\cite{dave} combines detection with verification, and PseCo~\cite{pseco} connects point localization, segmentation, and counting.
CountGD~\cite{countgd} and CountGD++~\cite{countgdpp} adapt open-vocabulary detection foundations and develop flexible target specification for open-world counting.
CountVid~\cite{countvid} extends this line to videos by linking instance predictions over time.
These advances strengthen the connection between user-specified targets, visual evidence, and instance enumeration.

Recent work further adapts foundation models to difficult counting conditions.
SAM3Count~\cite{sam3count} uses a density-aware image-counting pipeline with adaptive region tiling to improve dense-target recovery.
AdaCount~\cite{adacount} uses prototype-based similarity to guide spatial warping and feature modulation, reallocating representational capacity toward relevant targets without counting-specific retraining.
These studies demonstrate the value of adapting spatial processing to target density and scale.
Spatial grounding is also becoming an explicit capability of generalist vision-language models.
MolmoPoint~\cite{molmopoint} develops structured pointing through grounding tokens, GroundAnything~\cite{groundanything} investigates efficient parallel visual grounding, and WeVisOne-VL~\cite{wevisone} unifies object-level and pixel-level perception.
Their progress provides a broader foundation for producing localized visual predictions in response to user-specified targets.

However, jointly maintaining semantic target selection and reliable instance enumeration across heterogeneous domains, object scales, and densities remains challenging.
Count Anything addresses this challenge by jointly training target-conditioned region and dense point predictors under a shared cross-domain counting task.
Point-centric supervision enables joint learning from point-only and box-annotated instances, and complementary fusion combines their predictions into a single instance set.
This design connects semantic target specification, object-level spatial structure, and dense-instance localization within the same counting model.

\subsection{Counting Datasets and Evaluation}

Domain-specific datasets establish the empirical foundations of object counting.
ShanghaiTech~\cite{shanghaitech}, JHU-CROWD++~\cite{jhu_crowd}, and NWPU-CROWD~\cite{nwpu_crowd} support crowd counting over substantial variations in density and appearance.
Datasets for vehicles, crops, cell nuclei, and microorganisms extend counting to additional imaging conditions and object types~\cite{carpk,gwhd,bcdata,agar}.
These resources provide detailed instance supervision and expose the diverse scales, densities, and visual structures encountered in practical counting applications.

Category-diverse benchmarks expand the range of targets available for training and evaluation.
FSC-147~\cite{fsc147} supports category-agnostic counting, FSCD-LVIS~\cite{fscd_lvis,lvis} connects counting with detection across a larger category space, and OmniCount-191~\cite{omnicount191} develops multi-label counting evaluation.
Recent studies further examine whether predictions follow the requested target semantics.
PrACo++~\cite{praco_pp} introduces negative-label and distractor tests, together with MUCCA, a collection of real-world images containing multiple annotated categories.
These evaluations show the importance of checking target selection alongside numerical counting accuracy.

Synthetic data generation provides an additional route to scaling counting supervision.
KubriCount~\cite{kubricount} combines controllable scene synthesis, image editing, and filtering to construct instance-annotated data across several semantic granularity levels; its associated HieraCount model studies counting under these target definitions.
MixCount~\cite{mixcount} generates mixed-object scenes with precise instance annotations and detailed target descriptions, addressing difficult combinations of similar objects, self-similar structures, and background distractors.
These efforts improve control over scene composition, annotation quality, and semantic counting targets, expanding the resources available for training and evaluating generalist counters.

Large detection, segmentation, grounding, and specialized-domain datasets provide further opportunities to extend counting coverage~\cite{mscoco,objects365,lvis,dota_cvpr,xview,cityscapes,monusac,livecell,soybean_pod_uav}.
However, differences in category definitions, countable-instance criteria, and annotation types complicate their direct use for unified cross-domain counting.
CLOC develops this direction across six visual domains by organizing heterogeneous annotations into counting points with reliable bounding boxes retained when available.
Its large-scale coverage and consistent counting semantics support joint learning across categories, imaging conditions, object scales, and density distributions.
Together, CLOC and Count Anything connect the broad coverage of multi-source data with instance-grounded prediction in a shared training and evaluation framework.

\FloatBarrier

\end{document}